\documentclass[10pt,journal,compsoc]{IEEEtran}

\ifCLASSOPTIONcompsoc
  \usepackage[nocompress]{cite}
\else
  \usepackage{cite}
\fi
\ifCLASSINFOpdf
\else
\fi
\usepackage{graphicx}
\usepackage{multirow}
\usepackage{times}
\usepackage{amsmath,amsthm}
\usepackage{amssymb}
\usepackage{amsfonts}
\usepackage{algorithm}
\usepackage{mathrsfs}
\usepackage{booktabs} 
\usepackage{color}
\usepackage{hyperref} 
\usepackage{url}
\usepackage{algpseudocode}
\usepackage{pifont}
\usepackage{bm}
\usepackage{comment}
\usepackage{wasysym}
\usepackage{bbding-zy}

\def\1{{\bf{1}}}
\def\0{{\bf{0}}}

\def\a{{\bf a}}

\def\h{{\bf h}}

\def\q{{\bf q}}

\def\u{{\bf u}}

\def\z{{\bf z}}

\def\H{{\bf H}}

\def\W{{\bf W}}

\def\Bcal{{\mathcal{B}}}

\def\Gcal{{\mathcal{G}}}
\def\Hcal{{\mathcal{H}}}

\def\Lcal{{\mathcal{L}}}
\def\Mcal{{\mathcal{M}}}
\def\Ncal{{\mathcal{N}}}

\def\Rcal{{\mathcal{R}}}

\def\Tcal{{\mathcal{T}}}

\def\Vcal{{\mathcal{V}}}

\def\Ycal{{\mathcal{Y}}}

\def\Rbb{{\mathbb R}}

\newtheorem{definition}{Definition}

\begin{document}
%
\title{Learning Bi-typed Multi-relational Heterogeneous Graph via Dual Hierarchical Attention Networks}
%
%
%
%

\author{Yu~Zhao,
        Shaopeng~Wei,
        Huaming~Du,
        Xingyan~Chen,
        Qing~Li,~\IEEEmembership{Member,~IEEE},
        Fuzhen~Zhuang,~\IEEEmembership{Member,~IEEE},
        Ji~Liu~\IEEEmembership{Member,~IEEE},
        Gang~Kou~\IEEEmembership{Member,~IEEE}
\IEEEcompsocitemizethanks{
\IEEEcompsocthanksitem Y. Zhao, Q. Li and X. Chen are with Financial Intelligence and Financial Engineering Key Laboratory of Sichuan Province, Department of Artificial Intelligence, Southwestern University of Finance and Economics, China.\protect\\
E-mail: zhaoyu@swufe.edu.cn

\IEEEcompsocthanksitem S. Wei, H. Du and G. Kou are with School of Business Administration, Faculty of Business Administration, Southwestern University of Finance and Economics, Chengdu, 611130, China.
\protect\\
E-mail: kougang@swufe.edu.cn 

\IEEEcompsocthanksitem F. Zhuang is with Institute of Artificial Intelligence, Beihang University, Beijing 100191, China; and with 
SKLSDE, School of Computer Science, Beihang University, Beijing 100191, China \protect\\
E-mail:zhuangfuzhen@buaa.edu.cn 

\IEEEcompsocthanksitem J. Liu is with Kuaishou Technology, USA.\protect\\
E-mail:ji.liu.uwisc@gmail.com
\IEEEcompsocthanksitem G. Kou is the corresponding author.

}
}

\IEEEtitleabstractindextext{%
\begin{abstract}
Bi-type multi-relational heterogeneous graph (BMHG) is one of the most common graphs in practice, for example, academic networks, e-commerce user behavior graph and enterprise knowledge graph. It is a critical and challenge problem on how to learn the numerical representation for each node to characterize subtle structures. 
However, most previous studies treat all node relations in BMHG as the same class of relation without distinguishing the different characteristics between the intra-class relations and inter-class relations of the bi-typed nodes, causing the loss of significant structure information. 
To address this issue, we propose a novel \textbf{D}ual \textbf{H}ierarchical \textbf{A}ttention \textbf{N}etworks (DHAN) based on the bi-typed multi-relational heterogeneous graphs to learn comprehensive node representations with the intra-class and inter-class attention-based encoder under a hierarchical mechanism.
%
Specifically, the former encoder aggregates information from the same type of nodes, while the latter aggregates node representations from its different types of neighbors. Moreover, to sufficiently model node multi-relational information in BMHG, we adopt a newly proposed hierarchical mechanism. 
By doing so, the proposed dual hierarchical attention operations enable our model to fully capture the complex structures of the bi-typed multi-relational heterogeneous graphs. 
Experimental results on various tasks against the state-of-the-arts sufficiently confirm the capability of DHAN in learning node representations on the BMHGs.

\end{abstract}

\begin{IEEEkeywords}
Bi-typed Multi-relational Heterogeneous Graph, Graph Learning, Dual Hierarchical Attention Networks, Graph Neural Networks
\end{IEEEkeywords}}

\maketitle

\IEEEdisplaynontitleabstractindextext

%
\IEEEpeerreviewmaketitle

\begin{figure}
    \centering
    \includegraphics[width=0.47\textwidth]{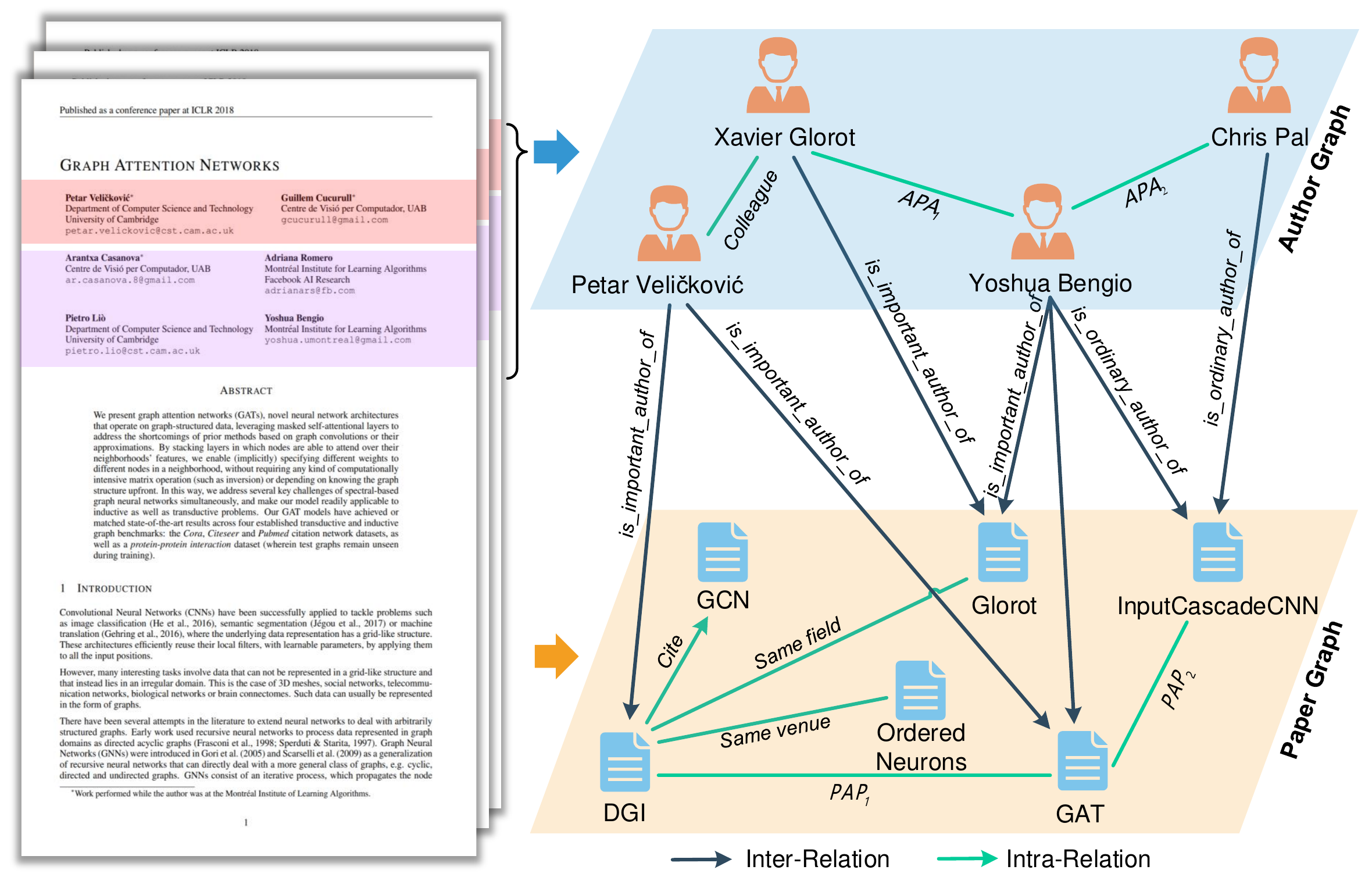}
    \caption{A toy example of {bi-typed multi-relational heterogeneous graph} (BMHG) in the academic networks. This graph consists of two types of nodes, i.e. \textit{authors} and \textit{papers}. Their links could be divided into two classes: (1) node intra-class relations, such as "\textit{colleague}" between authors, "\textit{cite}" and "\textit{same venue}" between papers; (2) node inter-class relations, such as "\textit{is\_important\_author\_of}", "\textit{is\_ordinary\_author\_of}". Table \ref{table-data-statistics} reports a detailed statistics of the graph.}
    \label{figure-oag-example}
\end{figure}

\ifCLASSOPTIONcompsoc
\IEEEraisesectionheading{\section{Introduction}\label{sec:introduction}}
\else

\section{Introduction}
\label{sec:introduction}
\fi
\IEEEPARstart{B}{i-typed} multi-relational heterogeneous graph (BMHG) typically consists of two different types of nodes and multiple intra-class/inter-class relations among them, which are ubiquitous in the real-world scenarios \cite{Wang2020A}, such as academic social networks \cite{Wang2012Influence,Zhang2019Heterogeneous}, 
e-commerce user behavior graph \cite{Xu2019Relation-Aware,niu2020dual}, and enterprise knowledge graph \cite{guleva2015using,li2020systemic}. These graphs have rich and valuable heterogeneous information that is worth deep mining.
For more clarity, we formally define the BMHG in Definition \ref{definition-HRG}. 
Without loss of generality, let us take OAG dataset \cite{Zhang2019Heterogeneous} as an example of the BMHG, which consists of two types of nodes, i.e. \textit{authors} and \textit{papers}, and multiple relationships, i.e. \textit{colleague}, \textit{cite}, \textit{is\_ordinary\_author\_of}, etc, as shown in Figure \ref{figure-oag-example}. 

\begin{table*}[htb]
    \begin{center}
         \caption{Comparison between several SOTA methods and the proposed model in terms of nodes heterogeneity and edges heterogeneity.}
         \label{Table-heterogeneity-statistics}
         \newcommand{\tabincell}[2]{\begin{tabular}{@{}#1@{}}#2\end{tabular}}
        \resizebox{0.95\textwidth}{!}{
           \begin{tabular}{l|l|c|c|c}
            \toprule
            \multicolumn{2}{c|}{\textbf{Models}} &
            \multicolumn{3}{c}{\textbf{Graph Heterogeneity}}    \\
            \midrule
            \textbf{Name} &  \textbf{Main ideas}& \tabincell{c}{\textbf{Bi-typed} \\
             $|\Tcal_{}|=2$} &\tabincell{c}{\textbf{Inter-class multi-relations}\\ $|\Rcal_\text{inter}|>1$ }&
            \tabincell{c}{\textbf{Intra-class multi-relations} \\$|\Rcal_\text{intra}|>1$}         \\
            \midrule
            \textbf{GCN\cite{Kipf2017Semi-supervised}}& \tabincell{c}{$\bullet$ Average  message passing } & \XSolidBrush & \XSolidBrush&\XSolidBrush \\
            \midrule
            \textbf{GAT\cite{Velickovic2018Graph}} &\tabincell{c}{$\bullet$ Attention based message passing} & \XSolidBrush&\XSolidBrush &\XSolidBrush \\
            \midrule
            \textbf{RGCN \cite{Schlichtkrull2018Modeling}} & \tabincell{l}{$\bullet$ Multi-relations \\ $\bullet$ Hierarchical weighting message passing } &\XSolidBrush &\XSolidBrush &\Checkmark\\
            \midrule
            \textbf{GTN \cite{Yun2019Graph}} & \tabincell{l}{$\bullet$ Multi-relations \\ $\bullet$ Self-adaption weighting message passing }&
             \XSolidBrush&\Checkmark& \XSolidBrush \\
            \midrule
            \textbf{HAN \cite{Wang2019Heterogeneous}}& \tabincell{l}{$\bullet$ Meta-path relations\\ $\bullet$ Hierarchical attention based message passing} &
             \XSolidBrush&\Checkmark& \XSolidBrush \\
            \midrule
            \textbf{HetGNN \cite{Zhang2019Heterogeneous}}&\tabincell{c}{$\bullet$ Heterogeneous features of nodes \\ $\bullet$ Attention based message passing } &
              \Checkmark &\XSolidBrush &\XSolidBrush\\
            
            \midrule
            \textbf{HGT \cite{Hu2020Heterogeneous}}  & \tabincell{l}{$\bullet$ Self-attention based message passing\\ $\bullet$ Node-balanced graph sampling\\ $\bullet$ Time  encoding } &\XSolidBrush & \Checkmark&\Checkmark \\
            \midrule
            \textbf{HGConv \cite{Yu2020Hybrid}} & \tabincell{l}{$\bullet$ Multi-relations \\ $\bullet$ Hierarchical attention based message passing}   & \XSolidBrush& \Checkmark& \XSolidBrush \\
            
            \midrule
            \textbf{ie-HGCN} \cite{Yang2021Interpretable} & \tabincell{l}{$\bullet$ Meta-path based relations \\ $\bullet$ Hierarchical attention based message passing} &\XSolidBrush &\Checkmark &\XSolidBrush  \\
            
            \midrule
            
            \textbf{DHAN (Ours)} &\tabincell{l}{$\bullet$ Distinguish inter-class relationship and intra-class relationship \\ $\bullet$ A newly proposed global-local hierarchical mechanism}  & \Checkmark&\Checkmark& \Checkmark  \\
            \bottomrule
           \end{tabular}
         }
    \end{center}
\end{table*}

\begin{definition}
\label{definition-HRG}
\textbf{Bi-typed Multi-relational Heterogeneous Graph}. 
A bi-typed multi-relational heterogeneous graph is defined as a connected graph $\Bcal\Mcal\Hcal\Gcal=(\Vcal,\Lcal,\Tcal,\Rcal)$. $\Vcal$ denotes the node set, and $\Lcal$ denotes a link set. 
They are associated with two functions: (i) a node type mapping function $\varphi: \Vcal \to \Tcal$, $|\Tcal|=2$. $\Vcal = \{\Vcal_1, \Vcal_2\}$, $\Vcal_1 \cap \Vcal_2 = \emptyset$. Each node $v \in \Vcal$ belongs to one particular node type in the node type set $\Tcal: \varphi(v) \in \Tcal$. 
(ii) a link class mapping function $\psi:\Lcal \to \Rcal$
. $\forall l_1, l_2 \in \Lcal$, $\psi(l_1) \in \Rcal_\text{intra}$ and $\psi(l_2) \in \Rcal_\text{inter}$ denote the node intra-class relationships and the node inter-class relationships, respectively.
$\Bcal\Mcal\Hcal\Gcal$ has multiple relationships, i.e., $|\Rcal_\text{inter}| > |\Tcal|-1 >0$ and $|\Rcal_\text{intra}|>1$. 
\end{definition}

In this paper, we focus on how to encode the bi-typed multi-relational heterogeneous graphs, providing an effective and flexible way to use their structural knowledge. 
The ultimate goal is to pursue perfect low-dimension distributed representations for nodes and relations mainly according to heterogeneous information in BMHG.
The learned results are essential for the inference tasks over graph, such as link prediction \cite{Abu-El-Haija2018Watch,Feng2016GAKE}, node classification \cite{Lee2019Graph,Zhang2020Adaptive}, node clustering \cite{Wang2020A}, and graph classification \cite{Wu2019DEMO-Net,Lee2018Graph}. 

Previous heterogeneous graph learning studies attempt to adopt the advanced Graph Neural Networks (GNNs) to learn heterogeneous graph while preserving the heterogeneous structures \cite{Schlichtkrull2018Modeling,Wang2019Heterogeneous,Zhang2019Heterogeneous,Hu2020Heterogeneous}.
However, most previous studies treat all node relations in BMHG as the same class of relation without distinguishing the different characteristics between the intra-class relations and inter-class relations of the bi-typed nodes,
which inevitably leads to graph significant structural information loss.
To address the issue, we propose a novel \textbf{D}ual \textbf{H}ierarchical \textbf{A}ttention \textbf{N}etworks (DHAN) to learn comprehensive node representations based on the bi-typed multi-relational heterogeneous graph with the intra-class and inter-class attention-based encoder under a hierarchical mechanism. 
Specifically, the former encoder model aggregates information from the same type of nodes (Section \ref{Section-intra-class}), while the latter encoder aggregates node representations from its different types of neighbors (Section \ref{Section-inter-class}). Moreover, to sufficiently learn node multi-relational information in BMHG, we adopt a newly proposed hierarchical mechanism. 
By doing so, the proposed dual hierarchical attention operations enable our model to fully capture the complex structures of the bi-typed multi-relational heterogeneous graphs. 
The comparison between previous existing methods with our proposed DHAN in terms of nodes heterogeneity and edges heterogeneity is shown in Table \ref{Table-heterogeneity-statistics}.

To evaluate the effectiveness of our proposed model, we generate three different kinds of datasets according to the popular Open Academic Graph (OAG) \cite{Zhang2019Heterogeneous} with various paper citation thresholds, including \textit{OAG1Y}, \textit{OAG2Y} and \textit{OAG10Y}. We conduct extensive experiments on these datasets with author disambiguation and paper classification task against the state-of-the-art methods, which sufficiently demonstrate the better capability of our proposed DHAN in learning node representations in the bi-typed multi-relational heterogeneous graphs.

The contributions of our work are summarized as follows: 
\begin{itemize}
\item In this paper, we focus on embedding the bi-typed multi-relational heterogeneous graphs. To the best of our knowledge, no one attempts to deal with the task before.
This paper is expected to further facilitate the bi-typed heterogeneous graph-involved applications, such as academic network mining \cite{Hu2020Heterogeneous}, recommendation system \cite{guo2021dual}, enterprise knowledge graph embedding \cite{gualdi2016statistically}, etc.
\item To tackle the bi-typed multi-relational heterogeneous graph learning task, we propose a novel dual hierarchical attention networks (DHAN). Specifically, we equipped DHAN with the intra-class and inter-class attention networks under a newly proposed hierarchical mechanism, which enables the proposed model to sufficiently capture the complex structural knowledge in the BMHG.
\item We conduct extensive experiments to evaluate the performance of the proposed model. The results demonstrate the superiority of the proposed model against the SOTA methods for learning node representations on bi-typed multi-relational heterogeneous graphs. 
The source code and data of this paper can be obtained from: \url{https://github.com/superweisp/DHAN2022}.
\end{itemize}


The rest of the paper is organized as follows. In Section \ref{section-related-work}, we summarize and compare the related work. Section \ref{section-method} introduces the details of DHAN. 
Extensive experiments are conducted to evaluate the effectiveness of the proposed model in Section \ref{section-experiments}. Finally, we conclude the paper in Section \ref{section-future_work}.

\section{Related Work}
\label{section-related-work}
\subsection{Graph Learning}
Recent years have witnessed a growing interest in developing graph learning algorithms  \cite{Tang2015Line,Perozzi2014Deepwalk,Grover2016Node2vec,gui2021pine} since most real-world data can be represented by graphs conveniently. Classical graph learning methods aim to reduce the dimension of graph data into low-dimensional representations, such as the linear method PCA \cite{Jolliffe2016Principal} and the non-linear method LLE \cite{Roweis2020Nonlinear}. Inspired by the basic idea from probabilistic language models such as skip-gram \cite{Guthrie2006A} and bag-of-words \cite{Zhang2010Understanding}, some random walk-based methods are proposed to learn node representations, such as DeepWalk \cite{Perozzi2014Deepwalk} and its advanced extension Node2Vec \cite{Grover2016Node2vec}. DeepWalk is a popular random walk-based graph learning method, which uses local information obtained from truncated random walks to learn node representations. There are also some matrix factorization-based methods for graph learning tasks \cite{Belkin2003Laplacian,Singh2008Relational}.
We refer the readers to \cite{Chen2020Graph} for more surveys on graph learning. 

\subsection{Graph Neural Networks}
Graph Neural Networks (GNNs) develop a deep neural network to deal with arbitrary graphs for representation learning \cite{Zhou2019Graph,Zhang2018Link,Hou2020Measuring,Hu2020Heterogeneous,Wang2020Heterogeneous}. 
GNNs have been successfully applied to various tasks over graphs \cite{Velickovic2018Graph,LEE2018Attention}, such as graph classification \cite{Wu2019DEMO-Net,Lee2018Graph}, link prediction \cite{Abu-El-Haija2018Watch}, and node classification \cite{Lee2019Graph,Zhang2020Adaptive}. The Graph Convolutional Networks (GCNs), as a representative GNN model, generalize convolutional operation on the graph-structured data \cite{Xu2019Relation-Aware,Schlichtkrull2018Modeling,Xu2019How,Vashishth2020Composition}. Graph Attention Networks (GATs) learn from the underlying graph structure by incorporating attention mechanism into GCNs \cite{Xu2019How,Vashishth2020Composition}, where the hidden representation of each node is computed by recursively aggregating and attending over its corresponding local neighbors’ features, and the weighting coefficients are calculated inductively with {self-attention} strategy \cite{Thekumparampil2018Attention-based,Qian2018Translating}. We refer the readers to \cite{Zhou2019Graph} for more references of GNNs. 
Despite the success of the above methods, they are constrained to perform only on homogeneous graphs, which thus could not handle the rich information in heterogeneous graphs.

\subsection{Heterogeneous Graph Neural Networks}
Heterogeneous graphs contain different types of nodes and edges \cite{Shi2017A,Wang2019Heterogeneous,Hu2019Heterogeneous,Zhang2019Heterogeneous}, which have rich and valuable heterogeneous information. Heterogeneous graph modeling methods are useful for various task, such as short text classification \cite{Hu2019Heterogeneous}, spam review detection \cite{Li2019Spam}, recommendation system \cite{Wang2019KGAT}, node and graph classification \cite{Wu2019DEMO-Net}, conversation generation \cite{Zhou2018Commonsense}, sentiment analysis \cite{Wang2020Relational}. 
To deal with heterogeneous graphs,
Wang et al.\cite{Wang2019Heterogeneous} proposed heterogeneous graph attention networks (HAN), which mainly concentrate on the different meta-paths. Zhang et al. \cite{Zhang2019Heterogeneous} proposed HetGNN that uses specialized Bi-LSTM to integrate the heterogeneous node attributes and neighbors. Schlichtkrull et al. \cite{Schlichtkrull2018Modeling} proposed RGCN to learn knowledge graphs \cite{Busbridge2019Relational} by employing relation-specific transformation matrices. Busbridge et al. \cite{Busbridge2019Relational} proposed RGAT by extending non-relational GATs to incorporate relational information, but with poor performance. 
Hu et al. \cite{Hu2020Heterogeneous} proposed heterogeneous graph transformer (HGT) to model web-scale heterogeneous graphs, which considers graph heterogeneity, dynamic nature and efficient training for large-scale graph. 

Despite their success, to the best of our knowledge, no one focuses on bi-typed multi-relational heterogeneous graph learning. 
Previous methods usually ignore the heterogeneous characteristics of inter-class and inter-class relationships of bi-typed nodes in BMHG.
Different from the conventional heterogeneous GNNs, this paper concentrates on the bi-typed heterogeneous graph learning task and attempts to design dual hierarchical graph attention networks to learn comprehensive node representations. Table \ref{Table-heterogeneity-statistics} summarizes the key advantages of our model in terms of modeling graph heterogeneity, compared with a variety of state-of-the-art heterogeneous GNNs models.

\section{Methodology}
\label{section-method}
\begin{figure*}
    \centering
    \includegraphics[width=0.9\textwidth]{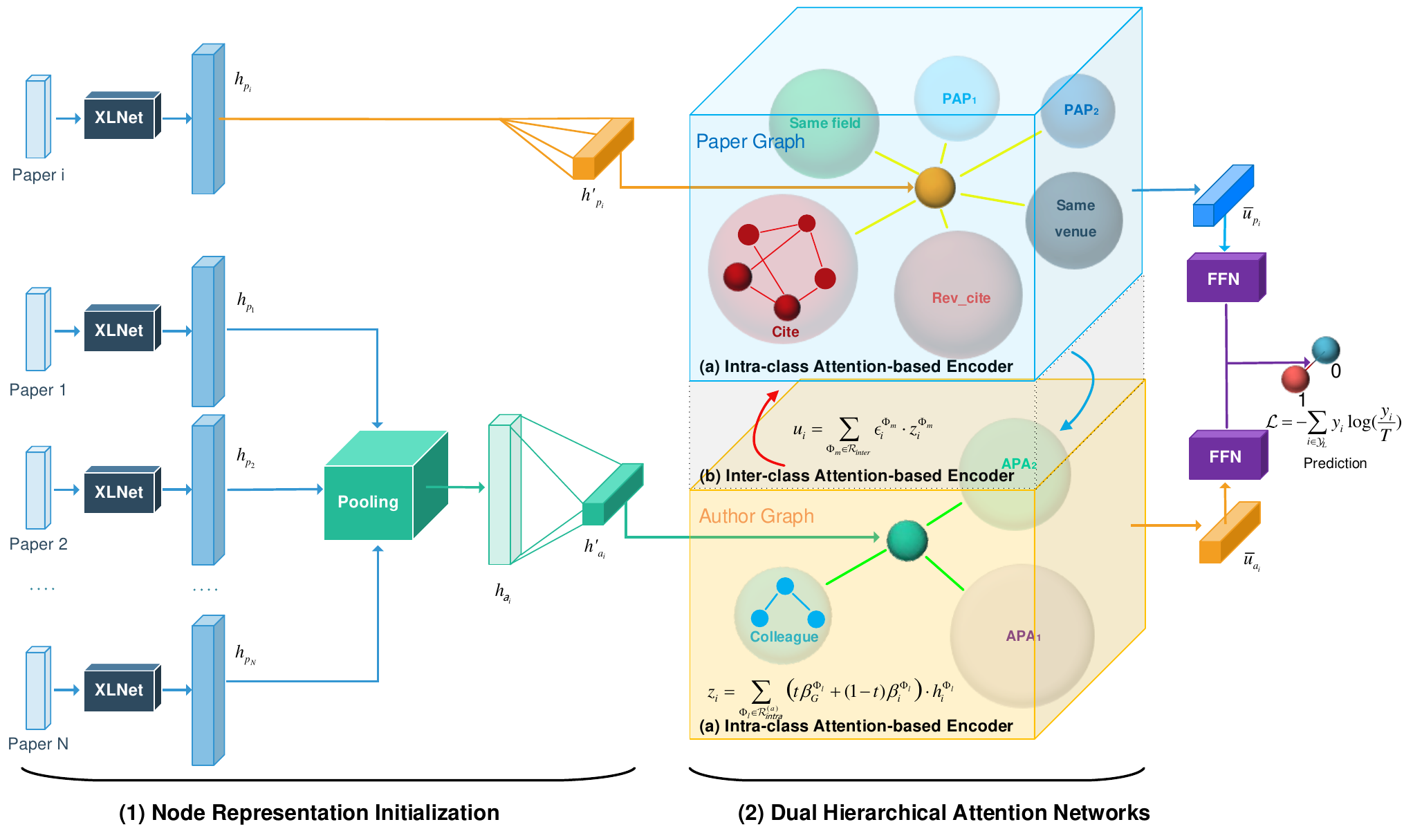}
    \caption{The overall architecture of the proposed method. 
    (1) \textbf{Node Representation Initialization} We utilize a pretrained XLNet to retrieval papers' representations from their titles. Authors' representations are calculated by averaging their published papers' information. (2) \textbf{Dual Hierarchical Attention Networks}: 
    (a) \textbf{Intra-class Attention-based Encoder} aims to aggregate different types of intra-class relationships with our novel hierarchical mechanism. For each type of node, the intra-class hierarchical attention module shares the same structure but with different parameters; (b) \textbf{Inter-class Attention-based Encoder} is designed for updating information between two types of nodes. Each type of node incorporates their inter-class neighbors' information with different weights according to different relationships and node pairs. 
    }
    \label{fig:overall-architecture}
\end{figure*}

\begin{figure}
    \centering
    \includegraphics[width=0.5\textwidth]{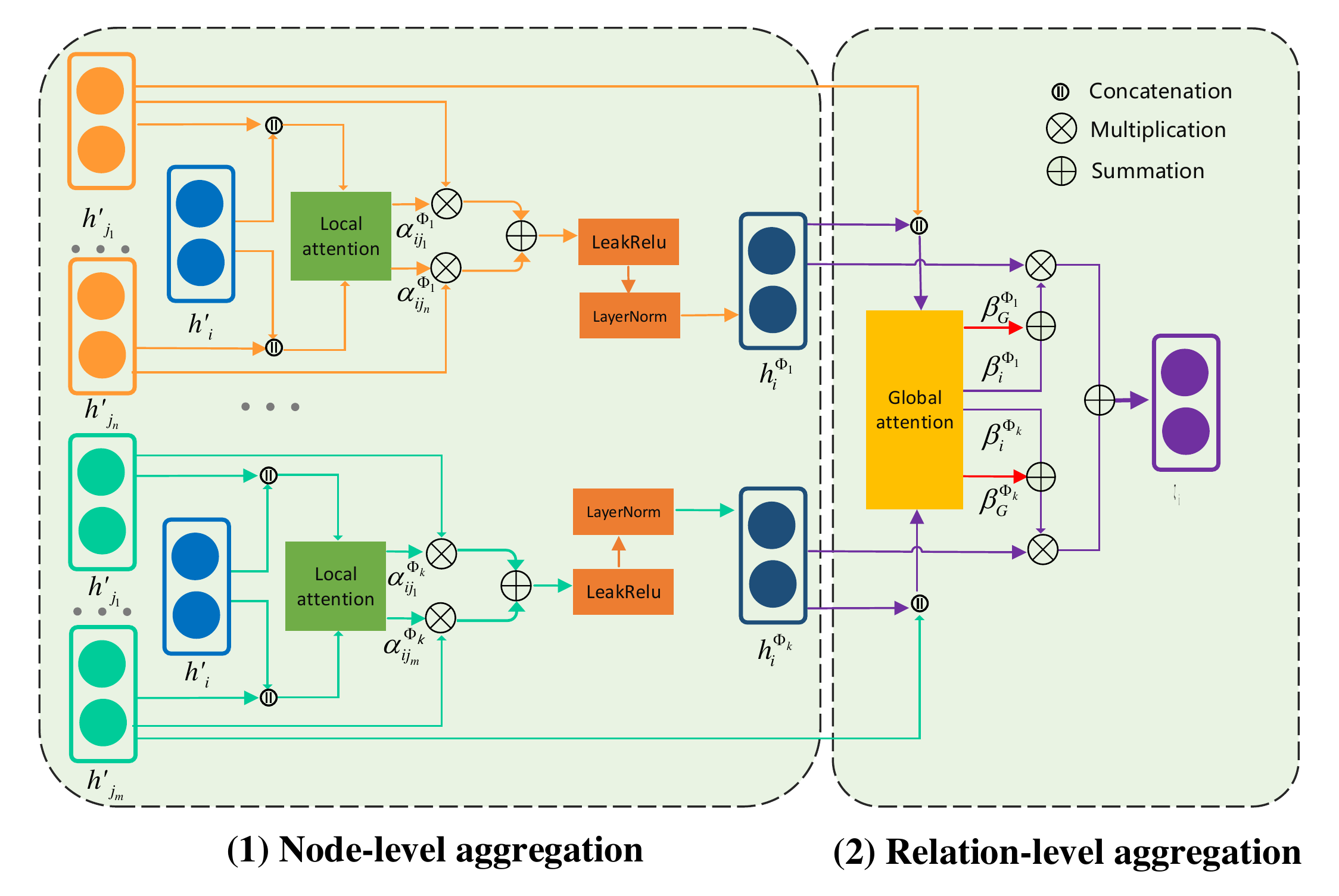}
    \caption{The proposed hierarchical attention mechanism. (1) Node-level aggregation aims to capture neighbor nodes' importance $\alpha_{ij}^{\Phi_k}$ under a specific type of relation $\Phi_k$. Then relation representations $\h_i^{\Phi_k}$ are aggregated by weighting and summing the target node's $\Phi_k$ based neighbors' information. (2) Relation-level aggregation  firstly assigns relation importance utilizing attention mechanism based on target node embedding and relation representations. Then relation representations are aggregated to get comprehensive neighbor information $\bm{\z}_i$ with final importance, which consists of global relation weight $\beta_G^{\Phi_l}$ and local relation weight $\beta_i^{\Phi_k}.$
    }.
    \label{fig:Hierarchical Mechanism}
\end{figure}

This section introduces the framework of the overall architecture, as shown in Figure \ref{fig:overall-architecture}. (1) Node Representation Initialization. We firstly initialize paper node representations through a pre-trained XLNet with their titles. Then we calculate author node representations by averaging their corresponding paper nodes' representations. (2) Dual Hierarchical Attention Networks (DHAN). The proposed DHAN consists of two submodules: intra-class attention-based encoder and inter-class attention-based encoder, which aim to fully capture the structural knowledge of BMHG. 
To model node multi-relational information in BMHG, we will introduce a newly proposed hierarchical mechanism, as shown in Figure \ref{fig:Hierarchical Mechanism}. Next, we gives the details of DHAN.


\subsection{Intra-class Attention-based Encoder}
\label{Section-intra-class}
The intra-class attention networks aim to learn the node embeddings by aggregating node information from their same type of neighbors, as shown in Figure \ref{fig:overall-architecture} (a).
Given a set of nodes with the same type $\Vcal_a \in \{\Vcal_1, \Vcal_2\}$, and a node pair ($v_i$, $v_j) (\in \Vcal_a$) that are connected via node intra-class relationship ${\Phi_k} \in \Rcal_\text{intra}^{(a)}$, we firstly perform transformation based on node type to project original node representation into $\Rbb^d$ latent space as follow:
\begin{equation}
\begin{array}{l}
\begin{aligned}
    \label{equation-projection-a}
     \H'^{(a)}=\W^{(a)}\H^{(a)},\\
\end{aligned}
\end{array}
\end{equation}
where $\W^{(a)} \in \Rbb^{d \times d'}$ are a trainable weight matrix related to a corresponding node type. $\H^{(a)} \in \Rbb^{| \Vcal_a | \times d}$ and $\H'^{(a)} \in \Rbb^{| \Vcal_a | \times d'}$ is the original and transformed node representations, respectively.


For node $v_i$, different types of intra-class relationships contribute different semantics to its embeddings, and so do different nodes with the same relationship. Hence, we then employ attention mechanism here in node-level and relation-level to hierarchically aggregate signals from the same types of neighbors to target node $v_i$.
We first perform {self-attention} on the nodes to formulate the importance $e_{ij}^{\Phi_k}$ of a specific-relation based node pair ($v_i$, $v_j$):

\begin{equation}
\begin{array}{l}
\begin{aligned}
    \label{equation-att-intra-class-node}
     e_{ij}^{\Phi_k}=  \text{att}_{\text{local}}(\h'_i,\h'_j;\Phi_k)
     = \text{LeakyRelu}(\a_{\Phi_k}^\top \cdot [ \h'_i \| \h'_j]) \ ,
\end{aligned}
\end{array}
\end{equation}
where $\h'_i \in \Rbb^{d'},\h'_j \in \Rbb^{d'}$ are transformed hidden representations of the node $v_i$ and $v_j$, respectively. $\|$ denotes the concatenate operation. 
$\a_{\Phi_k}^\top \in \Rbb^{2d'\times 1}$ is the shared node-level attention weight vector under relation $\Phi_k$. LeakyReLU is a nonlinearity activation function. 

Based on Eq. (\ref{equation-att-intra-class-node}), we calculate the $e_{ij}^{\Phi_k}$ for all nodes $v_j \in \Ncal_\text{intra}^{\Phi_k}(v_i)$, where $\Ncal_\text{intra}^{\Phi_k}(v_i)$ denotes specific relation-based neighbors of $v_i$. To make importance easily comparable across different nodes, we normalize them across all choices of $v_j$ using the softmax function:
\begin{equation}
    \alpha_{ij}^{\Phi_k} = \text{softmax}_j (e_{ij}^{\Phi_k}) = \frac{\exp{(e_{ij}^{\Phi_k}})}{\sum\limits_{v_p \in \Ncal_\text{intra}^{\Phi_k}(v_i) }\exp{(e_{ip}^{\Phi_k})}} \ ,
\end{equation}

Then, the embedding $\h_i^{\Phi_k}$ of node $v_i$ under given relation $\Phi_k$ is calculated by aggregating its intra-class neighbors' projected representations with the corresponding coefficients as follows:
\begin{equation}
    \label{equation-node-level-aggregation-intra-class}
    \h_i^{\Phi_k}  =  \text{LeakyRelu} \Big (\text{Norm}_{\Phi_k}\Big(\sum_{v_j \in \Ncal_\text{intra}^{\Phi_k}(v_i)}\alpha_{ij}^{\Phi_k} \cdot\h'_j \Big )\Big) \ ,
\end{equation}
where $\text{Norm}_{\Phi_k}$ denotes relation-specific layer normalization operation. Since the attention coefficient $\alpha_{ij}^{\Phi_k}$ is computed for a particular relationship, $\h_i^{\Phi_k}$ is semantic-specific and capable of capturing one kind of semantic information.


To learn more comprehensive node representations, we fuse different relation-specific aggregated information of nodes. 
Different from previous methods that either consider relation global importance \cite{Wang2019Heterogeneous} or local importance \cite{Yu2020Hybrid}, we take advantage of both of the two factors in relation-level attention, considering both the heterogeneity with regard to different nodes and the common information that a type of relation has among all nodes.
Firstly, we calculate the local importance $g_i^{\phi_k}$ of relation $\Phi_k$ with respect to node $v_i$ as follows:
\begin{equation}
   g_i^{\Phi_k}=\q^\top \Big(\h_i' \| \h_i^{\Phi_k} \Big),
\end{equation}
where $\q \in \Rbb^{2d' \times 1}$ is a trainable parameter. Then, we implement the softmax function to normalize the node-relation specific local importance across different relations.
\begin{equation}
    \beta_{i}^{\Phi_k} = \text{softmax}_k (g_{i}^{\Phi_k}) = \frac{\exp{(g_{i}^{\Phi_k})}}{\sum\limits_{\Phi_l \in \Rcal_{\text{intra} }^{(a)}}\exp{(g_{i}^{\Phi_l})}} \ ,
\end{equation}
where $\beta_{i}^{\Phi_k}$ indicates how important relation $\Phi_k$ is for node $v_i$, which measures local importance of intra-relation $\Phi_k$.
Secondly, to prevent model from local optimum and alleviate effects of noisy links, we design a relation global importance $\beta_G^{\Phi_l}$, which denotes how important intra-class $\Phi_l$ is for all nodes $v_i \in \Vcal_a $. 
Finally, as shown in \ref{fig:Hierarchical Mechanism}, we fuse different relation-specific aggregated information of nodes in both local and global view, as follow:
\begin{equation}
    \label{equation-global-local-aggregation}
    \bm{\z}_i  = \sum_{\Phi_l \in \Rcal_{\text{intra} }^{(a)}}\Big(t \beta_G^{\Phi_l}+(1-t) \beta_{i}^{\Phi_l } \Big)\cdot \h_i^{\Phi_l}  ,
\end{equation}
where $\bm{\z}_i \in \Rbb^{d'}$ is the learned representation of node $v_i$, which contains global and local information. $\h_i^{\Phi_l}$ denotes aggregated information for node $v_i$ under intra-class relation $\Phi_l$. $t$ is a smooth parameter to balance the global and local importance of intra-class relation $\Phi_l$. $\beta_G^{\Phi_l}$ and $t$ can be learned from training.

\subsection{Inter-class Attention-based Encoder}
\label{Section-inter-class}


Different from the above intra-class attention networks, the inter-class attention-based encoder aims to deal with the interaction between different types of nodes. We set $v_i^{(1)} \in \Vcal_1$ and $v_j^{(2)} \in \Vcal_2$. $\bm{\z}_i^{(1)},\bm{\z}_j^{(2)}$ are the learned representations of the node $v_i^{(1)}$ and $v_j^{(2)}$ by intra-class attention networks, respectively. 

We calculate the node-level importance $c_{ij}^{\Phi_m}$ for all nodes $v_j \in \Ncal_\text{inter}^{\Phi_m}(v_i)$, where $\Ncal_\text{inter}^{\Phi_m}(v_i)$ denotes the  neighbors of node $v_i$ under specific inter-relation $\Phi_m$. We normalize them across all choices of $v_j$ using the softmax function:
\begin{equation}
\begin{array}{l}
\begin{aligned}
    \label{equation-att-inter-class-node}
     c_{ij}^{\Phi_m}&=  att_{node}(\bm{\z}_i,\bm{\z}_j;\Phi_m)\\
     &= \text{LeakyRelu}(\a_{\Phi_m}^\top \cdot [\W^{(1)} \bm{\z}_i \| \W^{(2)} \bm{\z}_j]) \ ,
\end{aligned}
\end{array}
\end{equation}

\begin{equation}
    \label{equation-softmax-inter-class-node}
    \gamma_{ij}^{\Phi_m} =\text{softmax}_j (c_{ij}^{\Phi_m})= \frac{\exp{(c_{ij}^{\Phi_m}})}{\sum\limits_{v_k \in \Ncal_\text{inter}^{\Phi_m}(v_i) }\exp{(c_{ik}^{\Phi_m})}} \ ,
\end{equation}
where $\W^{(1)}, \W^{(2)} \in \Rbb^{d' \times d'}$ are two type-specific matrices to map their features $\z_{i},\z_j$ into a common space. 
$\a_{\Phi_m} \in \Rbb^{2d'}$ is a trainable weight vector. Then, as shown in Figure \ref{fig:overall-architecture}, the relation representation of node $v_i^{(1)}$ can be aggregated by its different types of neighbors' representations with the corresponding coefficients as follows:

\begin{equation}
    \label{equation-node-level-aggregation-inter-class}
    \bm{\z}_i^{\Phi_m}  = \text{LeakyRelu} \Big (\text{Norm}_{\Phi_m}\Big(\sum_{v_j \in \Ncal_\text{inter}^{\Phi_m}(v_i)}\gamma_{ij}^{\Phi_m} \W^{(2)}\bm{\z}_j \Big )\Big) \ ,
\end{equation}
where 
$\text{Norm}_{\Phi_m}$ indicates layer normalization operation related to the inter-class relation.

Similar to the above hierarchical attention, all relation representations are fused to get the final representations: 
\begin{equation}
   f_i^{\Phi_m}=\tilde{\q}^\top \Big(\bm{\z}_i \| \bm{\z}_i^{\Phi_m} \Big),
\end{equation}

\begin{equation}
    \epsilon_{i}^{\Phi_m} = \text{softmax}_m (f_{i}^{\Phi_m}) = \frac{\exp{(f_{i}^{\Phi_m})}}{\sum\limits_{\Phi_n \in \Rcal_{\text{inter} }}\exp{(f_{i}^{\Phi_n})}} \ ,
\end{equation}
where $\tilde{\q} \in \Rbb^{2d'}$ is a projection vector. $f_i^{\Phi_m}$ denotes the importance of relation embedding $\bm{\z}_i^{\Phi_m}$ related to node $v_i^{(1)}$. We apply the the softmax function to make relation importance comparable within inter-class relations. 
The representation $\u_i$ of node $v_i$ us obtained by fusing these relation-specific representations.
\begin{equation}
    \label{equation-relation-level-aggregation-inter-class}
    \u_i  = \sum_{\Phi_m \in \Rcal_{\text{inter}}} \epsilon_{i}^{\Phi_m }\cdot \bm{\z}_i^{\Phi_m}  ,
\end{equation}
where $\Rcal_{inter}$ indicates the set of relations among different types of nodes, i.e., node inter-class links. 

In inter-class hierarchical attention, the aggregation of different nodes' embedding is seamlessly integrated, and they are mingled and interactively affected each other in nature, as demonstrated in Figure \ref{fig:overall-architecture} (b). 
\subsection{Weighted Residual Connection}
For both intra-class encoder and inter-class encoder, we use weighted residual connection and layer normalization to alleviate over-smooth in practice. 
\begin{equation}
    \label{equation-residual-connection-intra-class}
    \bar{\bm{\z}}_i  = \text{Norm} \Big(\lambda \sigma(\bm{\z}_i) + (1-\lambda)\h_i \Big)  , 
\end{equation}
\begin{equation}
    \label{equation-residual-connection-inter-class}
    \bar{\u}_i  = \text{Norm} \Big(\tilde{\lambda} \sigma(\u_i) + (1-\tilde{\lambda})\bm{\z}_i \Big)  ,
\end{equation}
where $\lambda$ and $\tilde{\lambda}$ are hyperparameters.

\subsection{Optimization}
We train our model by minimizing the cross-entropy loss. {Inspired by  \cite{Wu2021Self}, we promote the training efficiency by adding Temperature $T$ in the learning.}
\begin{equation}
    \Lcal=-\sum\limits_{i \in\Ycal_{L}} y_i\log(\frac{\tilde{y_i}}{T}) \ ,
\end{equation}
where $\Ycal_{L}$ is the set of labeled nodes. $y_i$ and $\tilde{y_i}$ are the ground truth and the predicted label for node $i$, respectively.

\section{Experiments}
\label{section-experiments}






\subsection{Experimental Settings}
\subsubsection{Datasets}
We generate three different kinds of datasets by extracting different sub-graphs from the popular Open Academic Graph (OAG) dataset \cite{Zhang2019Heterogeneous} with {various paper citation thresholds}, including \textit{OAG1Y}, \textit{OAG2Y} and \textit{OAG10Y}. In \textit{OAG1Y}, we only retain the papers which are cited more than once a year. In \textit{OAG2Y} and \textit{OAG10Y}, we loose the time constraints to 2 years and 10 years, respectively. 
They contain two types of nodes, i.e., authors and papers, and several preliminary links including (author, \textit{colleague}, author), (author, \textit{is\_important\_author\_of}, paper), (author, \textit{is\_ordinary\_author\_of}, paper), (paper, \textit{cite}, paper). 
Note that the ``\textit{important}" authorship indicates an author is the first or second author of a paper, and the ``\textit{ordinary}" authorship indicates an author is not the important author of a paper. The basic statistics of all datasets are included in Table \ref{table-data-statistics}. 
The intra-class relations of authors include: \textit{colleague}, \textit{APA1} and \textit{APA2}. APA1 and APA2 indicate the co-authorship of important authors and ordinary authors, respectively. 
The intra-class relations of papers include: \textit{cite, {rev\_cite}, is\_same\_venue\_of, is\_same\_field\_of} . 
The inter-class relation between author and paper includes: {\textit{is\_important\_author\_of} and \textit{is\_ordinary\_author\_of}.}

\begin{table}[htb]
    \caption{Statistics of the datasets \textit{OAG1Y}, \textit{OAG2Y} and \textit{OAG10Y}, which are extracted from the popular Open Academic Graph \cite{Zhang2019Heterogeneous} with various citation thresholds.}
    \label{table-data-statistics}
    \centering
    \newcommand{\tabincell}[2]{\begin{tabular}{@{}#1@{}}#2\end{tabular}}
    \resizebox{0.5\textwidth}{!}{
    \begin{tabular}[t]{l|l|r|r|r}
    \toprule
     \multicolumn{2}{c|}{ \textbf{Datasets} } & \textit{\textbf{OAG1Y}} &\textit{\textbf{OAG2Y}}  & \textit{\textbf{OAG10Y}  } \\
                 \midrule
\multirow{2}{*}{ \tabincell{l}{\textbf{Bi-typed}\\ \textbf{nodes}}} &\textbf{\textit{\#Papers}}             & 494,051      & 825,234      & 1,564,109       \\
& \textbf{\textit{\#Authors}}           & 480,575      & 734,451      & 1,266,569       \\
\midrule
\multirow{3}{*}{ \tabincell{l}{\textbf{Author}\\ \textbf{intra-relations}}} & \textbf{\textit{\#Colleague}}        & 285,393,669   & 562,821,414   & 1,400,301,929    \\
&\textbf{\textit{\#APA1}}             & 369,973      & 600,344      & 1,074,851       \\
&\textbf{\textit{\#APA2}}             & 1,015,964     & 1,413,447     & 2,059,826       \\
\midrule
\multirow{5}{*}{\tabincell{l}{\textbf{Paper}\\ \textbf{intra-relations}}} & \textbf{\textit{\#Cite/Rev\_Cite}}             & 4,847,142     & 7,367,512     & 22,407,910      \\
& \textbf{\textit{\#Same Field}}       & 160,283,374,629 & 440,183,678,370 & 1,548,687,874,807   \\
& \textbf{\textit{\#Same Venue}}       & 273,272,355   & 619,484,732   & 1,929,963,113    \\
 & \textbf{\textit{\#PAP1}}             & 3,022,137     & 5,966,848     & 13,450,631      \\
& \textbf{\textit{\#PAP2}}             & 4,973,945     & 9,042,142     & 17,648,847      \\
\midrule
\multirow{2}{*}{\tabincell{l}{\textbf{Author-Paper}\\ \textbf{inter-relations}}}&\textbf{\textit{\#Important author}} & 800,061      & 1,306,953     & 2,372,890       \\
& \textbf{\textit{\#Ordinary author}}  & 661,250      & 1,019,506     & 1,687,184       \\
\midrule
\multicolumn{2}{c|}{ \textbf{\textit{Training data Period}}  }          & \multicolumn{3}{c}{2000 - 2015}             \\
\multicolumn{2}{c|}{\textbf{\textit{Validation data Period}}}            & \multicolumn{3}{c}{2015 - 2016}             \\
\multicolumn{2}{c|}{\textbf{\textit{Testing data Period}}}             & \multicolumn{3}{c}{2016 - 2019}            \\
        \bottomrule
    \end{tabular}}
\end{table}

\subsubsection{Baselines}
To demonstrate the effectiveness of our proposed model DHAN, we compare it with three types of SOTA baselines: (1) the homogeneous graph neural networks which do not consider multi-relationships between nodes, such as GCN, GAT; (2) the heterogeneous graph neural networks which take different relationships into consideration, such as RGCN, HGT; (3) the heterogeneous networks which implement a hierarchical mechanism to aggregate different kinds of relations in graphs, such as HAN, HGConv.

\textit{Homogeneous models:}
\begin{itemize}
    \item Graph Convolutional Networks (GCN) \cite{Kipf2017Semi-supervised,chen2020simple}: a popular model which simply averages neighboring nodes' representations in aggregation. 
    \item Graph Attention Networks (GAT) \cite{Velickovic2018Graph}: a recent model which takes attention mechanism to align different weights to neighbors during the information aggregating process. 
\end{itemize}

\textit{Heterogeneous models:}
\begin{itemize}
    \item Relational Graph Convolutional Networks (RGCN) \cite{Schlichtkrull2018Modeling}: an advanced extension of GCN, which takes relation information into consideration by giving different weights for difference relationships.
    \item Heterogeneous graph neural network (HetGNN)
    \cite{Zhang2019Heterogeneous}: a multi-modal heterogeneous graph model which utilizes Bi-LSTM to  process multi-moding information, then applies attention mechanism in heterogeneous information fusing.
    \item Graph Transformer Networks (GTN) 
    \cite{Yun2019Graph}: a novel heterogeneous graph neural network  based on GCN which updates adjacent matrix of different relations during the training process.
    \item Heterogeneous Graph Transformer (HGT) \cite{Hu2020Heterogeneous}: a state-of-the-art model which implements on heterogeneous graph with different types of nodes and multiple relations.
    
\end{itemize}

\textit{Hierarchical models:}
\begin{itemize}
    \item Heterogeneous Graph Attention Network (HAN)
    \cite{Wang2019Heterogeneous}: one of the earliest model which implements hierarchical attention on graph neural network based on meta-path.
    \item Heterogeneous Graph Convolution (HGConv)
    \cite{Yu2020Hybrid}: an efficient model which utilizes hierarchical mechanism based on different node types and different relations.
    \item interpretable and efficient Heterogeneous Graph Convolutional Network (ie-HGCN)
    \cite{Yang2021Interpretable}: a SOTA model which firstly implements object-level aggregation and then aggregates type-level information based on different meta-paths.
\end{itemize}


\subsubsection{Model Setting and Training Details}
\label{training-protocol}
We implement DHAN with PyTorch and PyTorch Geometric (PyG). 
We use a pre-trained XLNet \cite{yang2019xlnet} to initialize the paper nodes' representations. Then the author nodes' initial representations are aggregated by averaging their published papers' embeddings. 
We set the dropout rate of DHAN among \{0.1, 0.2, 0.3, 0.4, 0.5\} and the temperature T from \{0.01, 0.05, 0.1, 1, 1.5, 10\}. The ${\ell_2}$ regularization weight is set from \{1e-4, 1e-3, 1e-2, 1e-1\}. 
For the paper field L1 task (PF\_L1), we add one more weighted residual connection in inter-class aggregation process without adding any new parameters.
All models are trained with AdamW optimizer with the Cosine Annealing Learning Rate Scheduler. 
For all the baseline models and DHAN, we use 128 hidden dimension. 
For each model, we run 200 epochs and choose the best which has higher NDCG and lower loss compared with former training processes on validation datasets in order to alleviate the overfitting problem. To obtain the experimental results of all baselines, we run official codes provided by the original papers.
Finally, we report the results of each model on the testing datasets.

\begin{table*}[h]\scriptsize
    \begin{center}
         \caption{
         \textbf{Classification and link prediction results.} Evaluation of different methods on three datasets.}
         \label{table-results}
         \newcommand{\tabincell}[2]{\begin{tabular}{@{}#1@{}}#2\end{tabular}}
        \resizebox{\textwidth}{!}{
           \begin{tabular}{c|c|c|ccccccccc|c}
            \toprule
            \textbf{Datasets} & \textbf{Tasks}  & \textbf{Metrics} &\textbf{GCN \cite{Kipf2017Semi-supervised}}  & \textbf{GAT \cite{Velickovic2018Graph}} & \textbf{RGCN \cite{Schlichtkrull2018Modeling}} &
            \textbf{HAN \cite{Wang2019Heterogeneous}} &
            \textbf{HetGNN \cite{Zhang2019Heterogeneous}}&
            \textbf{GTN \cite{Yun2019Graph}}&
            \textbf{HGT \cite{Hu2020Heterogeneous}} &
            \textbf{HGConv \cite{Yu2020Hybrid}} & 
            \textbf{ie-HGCN \cite{Yang2021Interpretable}}&
            \textbf{DHAN}  \\
            \midrule

             \multirow{7}{*}{\textit{OAG1Y}} &
            \multirow{2}{*}{PV} & NDCG & 0.2661 & 0.2750 & 0.2693&0.2880 &0.2375 &0.2680 &0.2970 &0.2885 &0.2465 &\textbf{0.2995}\\
            \cmidrule{3-13}
            & & MRR & 0.1295 &0.1391  &0.1335 &0.1508 &0.1031 &0.1300 &0.1623 &0.1502 &0.1069  &\textbf{0.1643}\\
            \cmidrule{2-13}
            &\multirow{2}{*}{PF\_L1} & NDCG & 0.7180 & 0.7271 &0.7492 &0.7227 & 0.6587 & 0.7408 & 0.7515&0.7476& 0.7304 &\textbf{0.7532}  \\
            \cmidrule{3-13}
            &  &MRR & 0.6892 & 0.6905 &0.7220& 0.6916 & 0.6189 & 0.7088 &0.7169 &0.7179&0.6996& \textbf{0.7213}\\
            \cmidrule{2-13}
             &  \multirow{2}{*}{PF\_L2} & NDCG & 0.3598 &0.3678 &0.4191&0.3817&0.3059&0.3910&0.4502&0.4209 &0.3297&\textbf{0.4512}\\
            \cmidrule{3-13}
             &  & MRR & 0.3156 &0.3300 &0.4311 &0.3593&0.2183&0.3725&0.4958&0.4403&0.2528&\textbf{0.4960} \\
            \cmidrule{2-13}
            & \multirow{3}{*}{AD}&NDCG& 0.7297& 0.7915& 0.7820& 0.7497&0.6430 &0.7403
            &0.8037 &0.7715&0.7539&\textbf{0.8222}\\
            \cmidrule{3-13}
            & & MRR & 0.6436 & 0.7241 & 0.7120 &0.6693 &0.5309&0.6567&0.7403&0.6982&0.6749&\textbf{0.7651}\\
            \cmidrule{3-13}
            & & ACC & 0.4627& 0.5678 &0.5610&0.5002 &0.3240& 0.4800& 0.6040& 0.5406 &0.5106&\textbf{0.6394}\\
             \cmidrule{1-13} 
             
            \multirow{7}{*}{\textit{OAG2Y}} 
            &\multirow{2}{*}{PV}& NDCG & 0.2604 &0.2780 &0.2739 & 0.2899& 0.2465&0.2569 &0.2947 &0.2862 &0.1828 &\textbf{0.2969}\\
            \cmidrule{3-13}
            & & MRR &0.1282 & 0.1445& 0.1376& 0.1553&0.1137 &0.1200 &0.1616 &0.1496 &0.0502 &\textbf{0.1629}\\
            \cmidrule{2-13}
            & \multirow{2}{*}{PF\_L1}&NDCG & 0.7076& 0.7271 & 0.7410 &0.7384 &0.6614&0.7284 &0.7455 &0.7438&0.7195 &\textbf{0.7520} \\
            \cmidrule{3-13}
            & &MRR & 0.6838 & 0.6985 &0.7131 &0.7069 & 0.6282 &0.6905&0.7075&0.7139&0.6861&\textbf{0.7177}\\
            \cmidrule{2-13}
            & \multirow{2}{*}{PF\_L2} & NDCG &0.3651 & 0.3737 &0.4275 & 0.3882 &0.3075&0.4000&0.4544&0.4265&0.3383&\textbf{0.4558}\\
           \cmidrule{3-13}
            & &  MRR & 0.3226 & 0.3427 & 0.4429& 0.3629 &0.2179 &0.3955 &0.4916 & 0.4391 &0.2694&\textbf{0.4925}\\
            \cmidrule{2-13}
            &\multirow{3}{*}{AD} & NDCG &0.6726  & 0.7783 & 0.7841& 0.7509&0.6258&0.7167&0.8040 & 0.7797 &0.6718&\textbf{0.8332}\\
           \cmidrule{3-13}
            & & MRR &0.5698 & 0.7073 &0.7147 &0.6716 &0.5099 &0.6260 &0.7410 &0.7093&0.5688&\textbf{0.7796}\\
            \cmidrule{3-13}
            & & ACC &0.3769 & 0.5539 &0.5584 &0.5073 &0.2959&0.4330 &0.5959&0.5513&0.3599&\textbf{0.6554}\\
            \cmidrule{1-13}
            
            \multirow{9}{*}{\textit{OAG10Y}} & 
            \multirow{2}{*}{PV}& NDCG & 0.2604 &0.2718 &0.2739 & 0.2598& 0.2515&0.2317 &0.2801 &0.2655 &0.2405 &\textbf{0.2816} \\
            \cmidrule{3-13}
            &  & MRR& 0.1282 &0.1399 &0.1376 & 0.1225& 0.1196&0.0971 &0.1445 &0.1287 &0.1047 &\textbf{0.1476}  \\
            \cmidrule{2-13}
            & \multirow{2}{*}{PF\_L1}&NDCG & 0.7219& 0.7300 & 0.7520 & 0.7169 &0.6837 &0.7339 &\textbf{0.7550} &0.7489 &0.7222& 0.7530\\
            \cmidrule{3-13}
            & &MRR & 0.6902 & 0.6950 &\textbf{0.7266} &0.6834 
            &0.6554 &0.6953 &0.7196 &0.7188 &0.6899&  0.7197\\
            \cmidrule{2-13}
            & \multirow{2}{*}{PF\_L2} & NDCG &0.3595 & 0.3641 &0.4205 & 0.3768 &0.3125& 0.3892&0.3877
            &0.4189 &0.3342&
            \textbf{0.4556}\\
           \cmidrule{3-13}
            & &  MRR & 0.3081 & 0.3184 & 0.4196& 0.3385 & 0.2274 &0.3679 &0.3735 &0.4214& 0.2559&\textbf{0.4868}\\
            \cmidrule{2-13}
            &\multirow{3}{*}{AD} & NDCG &0.6042  & 0.7201 & 0.7685 & 0.7169 &0.5712 &0.6804 &0.7979
            &0.7659 &0.6477&\textbf{0.8343}\\
            \cmidrule{3-13}
            & & MRR &0.4841 & 0.6326 &0.6955 &0.6284
            & 0.4430 &0.5807 &0.7338 & 0.6923 &0.5394&\textbf{0.7828} \\
            \cmidrule{3-13}
            & & ACC &0.2862 & 0.4645 &0.5426 &0.4615 & 0.2476 &0.3920 &0.5973 &0.5471  &0.3479&\textbf{0.6799}\\

            \bottomrule
           \end{tabular}
         }
    \end{center}
\end{table*}

\subsection{Classification and Link Prediction}
\subsubsection{Evaluation Protocol}
We evaluate our model on three tasks, including author disambiguation (AD), paper-venue (PV), paper-field in L1 level (PF\_L1) classification and paper-field in L2 level (PF\_L2) classification. 
In the datasets, the fields of papers are divided into several hierarchical levels (such as Operating system/ file system), and lower level means more detailed categories. In other words, L2 (such as 'file system') has much more categories than L1 (such as operating system).
The author disambiguation task could be treated as a link prediction task which aims to predict the possible link between the same name and their associated papers. 
Both of the paper-venue and paper-field classifications are multi-classification problem. In paper-venue classification, each paper belongs to only one venue, while each paper may belong to several fields of  L1 level and L2 level in paper-field classification tasks. 
We adopt accuracy (ACC), Normalized Discounted Cumulative Gain(NDCG) and Mean Reciprocal Rank (MRR) as evaluation metrics. 

\begin{table*}[htb]
  \caption{\textbf{Node clustering results.}}
  \label{cluster-table}
  \centering
  \resizebox{\textwidth}{!}{
  \begin{tabular}{cccccccccccc }
    \toprule
            \textbf{Datasets}  & \textbf{Metrics} &\textbf{GCN \cite{Kipf2017Semi-supervised}}  & \textbf{GAT \cite{Velickovic2018Graph}} & \textbf{RGCN \cite{Schlichtkrull2018Modeling}} &
            \textbf{HAN \cite{Wang2019Heterogeneous}} &
            \textbf{HetGNN \cite{Zhang2019Heterogeneous}}&
            \textbf{GTN \cite{Yun2019Graph}}&
            \textbf{HGT \cite{Hu2020Heterogeneous}} &
            \textbf{HGConv \cite{Yu2020Hybrid}} & 
            \textbf{ie-HGCN \cite{Yang2021Interpretable}}&
            \textbf{DHAN}  \\
            \midrule
            
            \multirow{2}{*}{\textit{OAG1Y}}
            & ARI & 0.0340& 0.0286&0.0308 & 0.0337&0.0141 &0.0350 &0.0636 &0.0276 &0.0451 &\textbf{0.0728} \\
            \cmidrule{2-12}
             & NMI &0.6566 &0.6477 &0.6469 &0.6541 &0.6231 &0.6652 &0.6746 &0.6489 &0.6479 &\textbf{0.6764}  \\
             \cmidrule{1-12}
            \multirow{2}{*}{\textit{OAG2Y}}
            & ARI & 0.0156& 0.0194 &0.0109 &0.0225 &0.0120 &0.0176 & 0.0134&0.0313&0.0090&  \textbf{0.0474} \\
            \cmidrule{2-12}
             & NMI &0.6577 & 0.6646& 0.6571&0.6666 & 0.6302& 0.6773& 0.6740& 0.6678& 0.4758 &\textbf{0.7012}  \\
             \cmidrule{1-12}
             \multirow{2}{*}{\textit{OAG10Y}}
            & ARI &0.0152 &0.0286 &0.0124 &0.0079 &0.0135 &0.0369 &0.0318 &0.0337 & 0.0018& \textbf{0.0370} \\
            \cmidrule{2-12}
             & NMI &0.6422 & 0.6477& 0.6510& 0.6449& 0.6207&0.6706 & 0.6727& 0.6744& 0.6491 &\textbf{0.6818}  \\

    \bottomrule
  \end{tabular}
  }
\end{table*}

\subsubsection{Results and Analysis}
The experimental results of the proposed model and SOTA baselines are reported in Table \ref{table-results}. 
We can observe from Table \ref{table-results} that our proposed DHAN outperforms all the baselines on all tasks across most of metrics on all datasets. 
For instance, our model improves the ACC, NDCG and MRR of author disambiguation on \textit{OAG1Y} from 0.6477 to 0.8343, 0.5394 to 0.7828, and 0.3479 to 0.6799 respectively comparing to the state-of-the-art model ie-HGCN, which confirms the capability of DHAN in learning bi-typed multi-relational heterogeneous graph. 

\begin{figure*}[htb]
    \centering
    \includegraphics[width=0.33\textwidth]{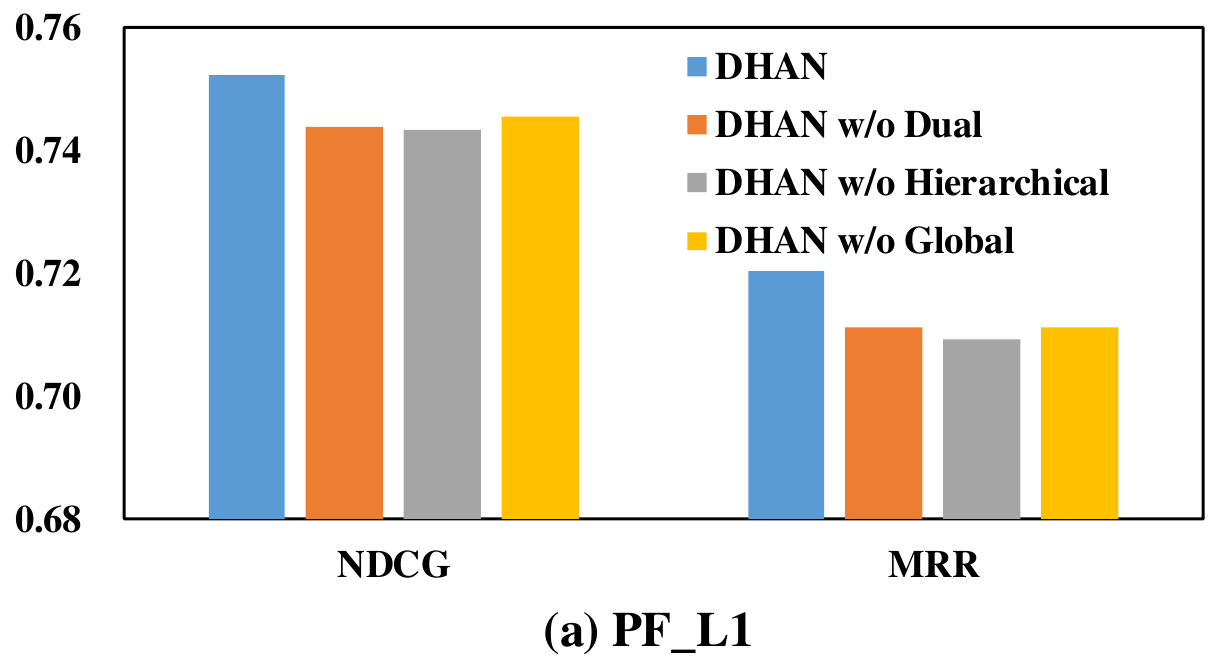}
    \includegraphics[width=0.33\textwidth]{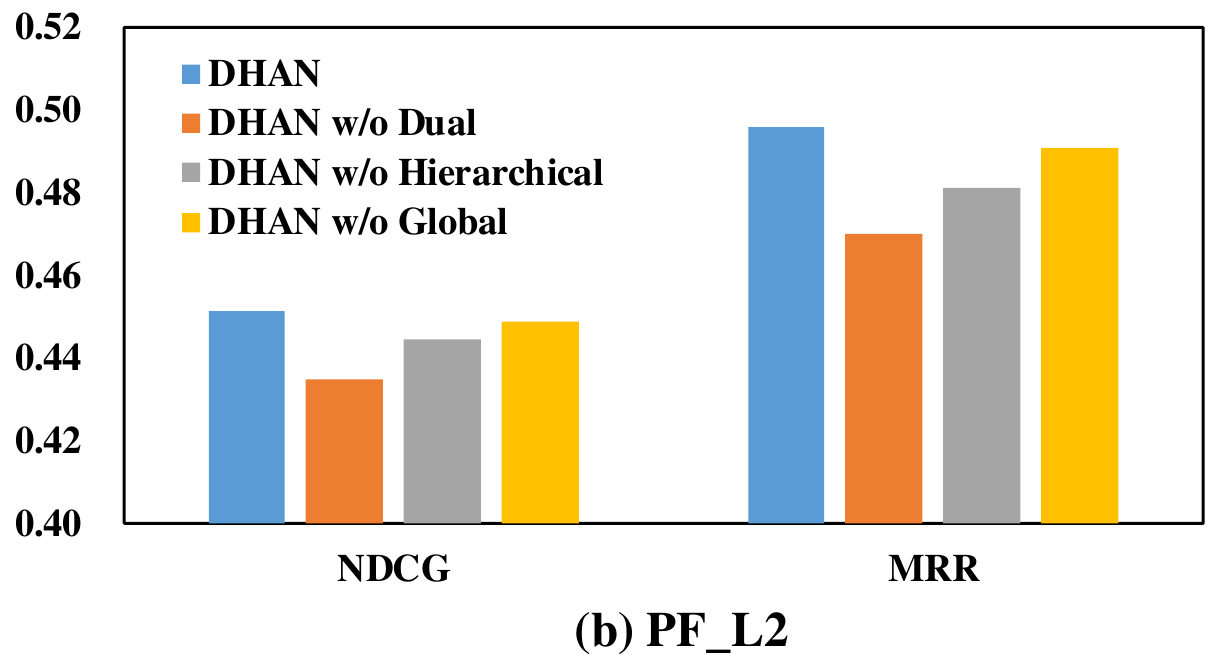}
    \includegraphics[width=0.33\textwidth]{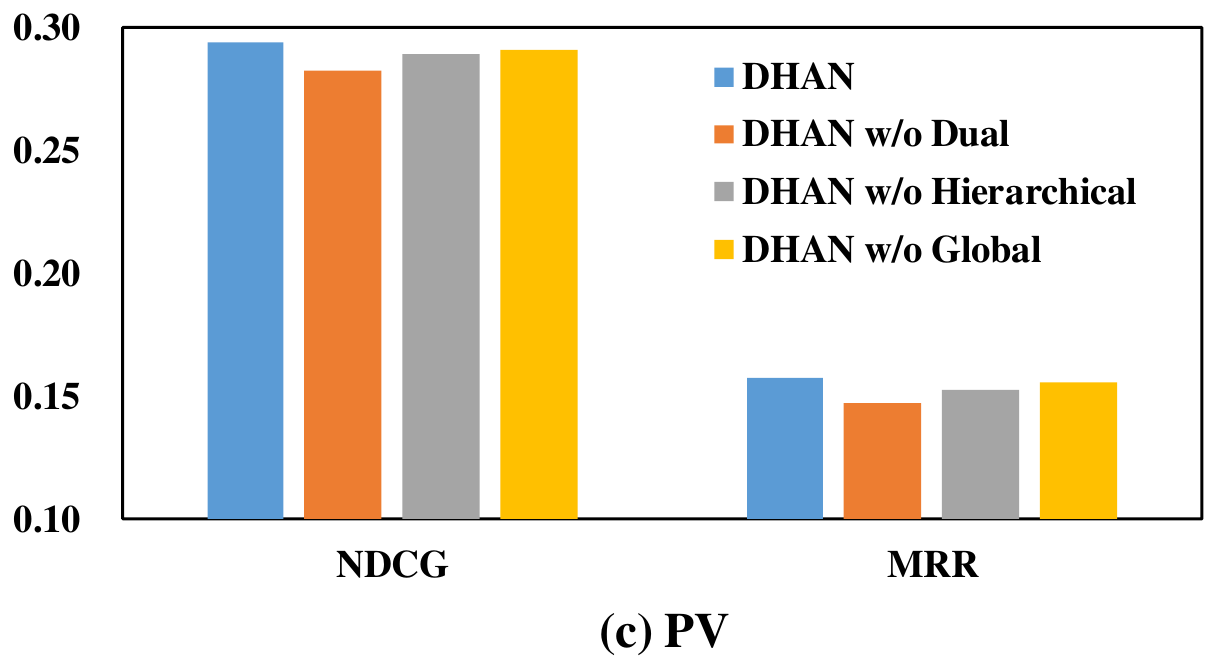}
    \caption{The ablation study of our model on \textit{OAG1Y}.}
    \label{fig:ablation-study}
\end{figure*}

\begin{figure*}[htb]
    \centering
    \includegraphics[width=0.16\textwidth]{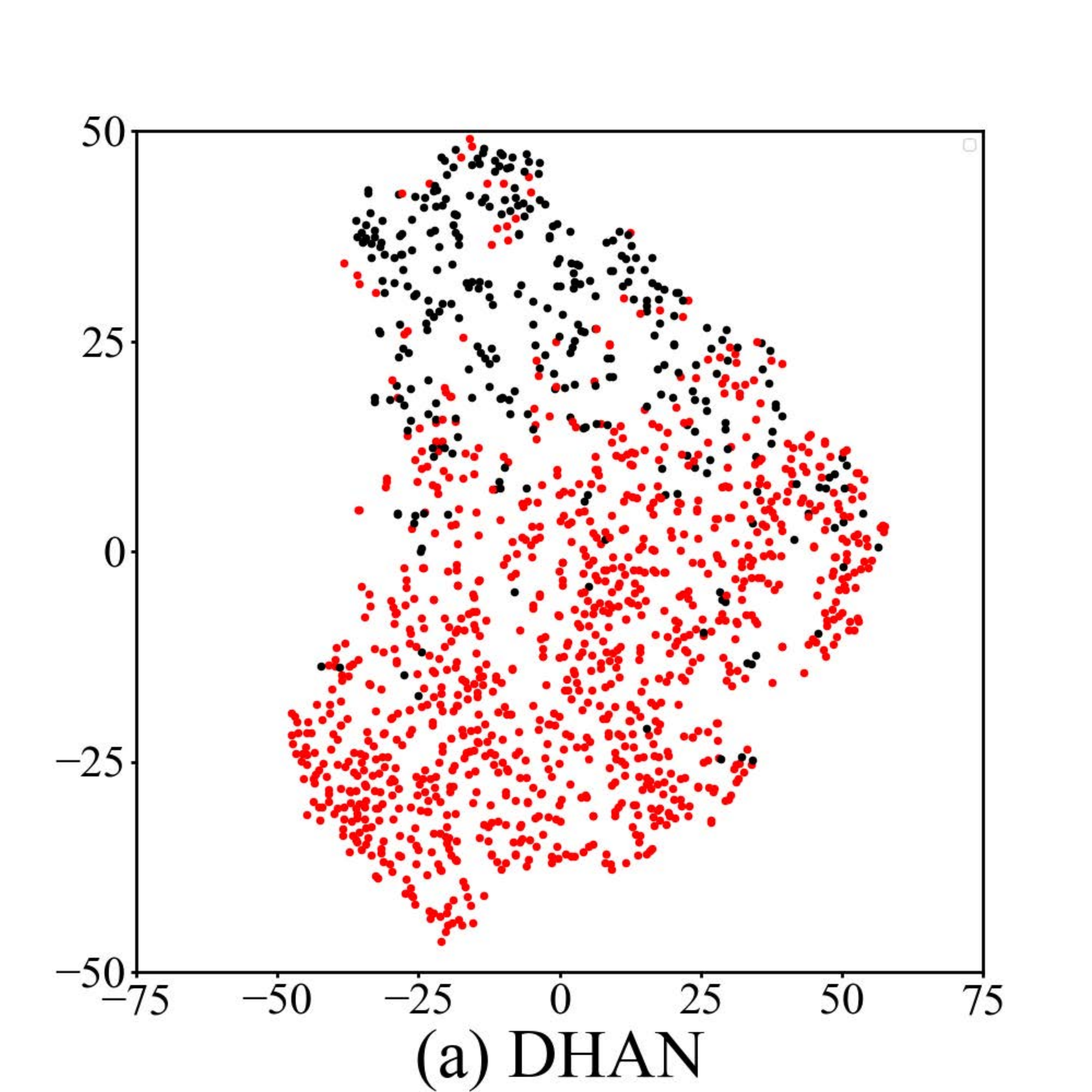}
    \includegraphics[width=0.16\textwidth]{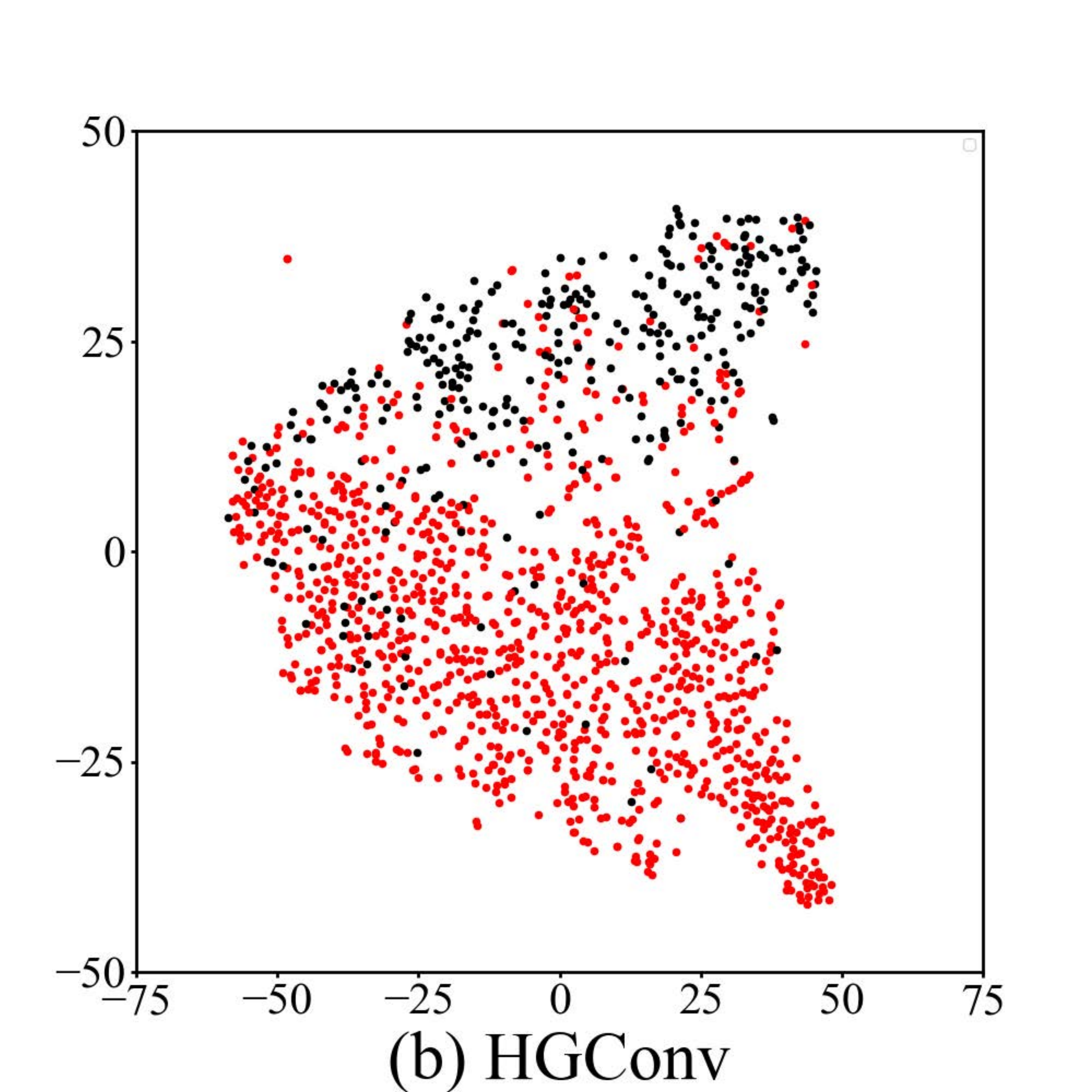}
     \includegraphics[width=0.16\textwidth]{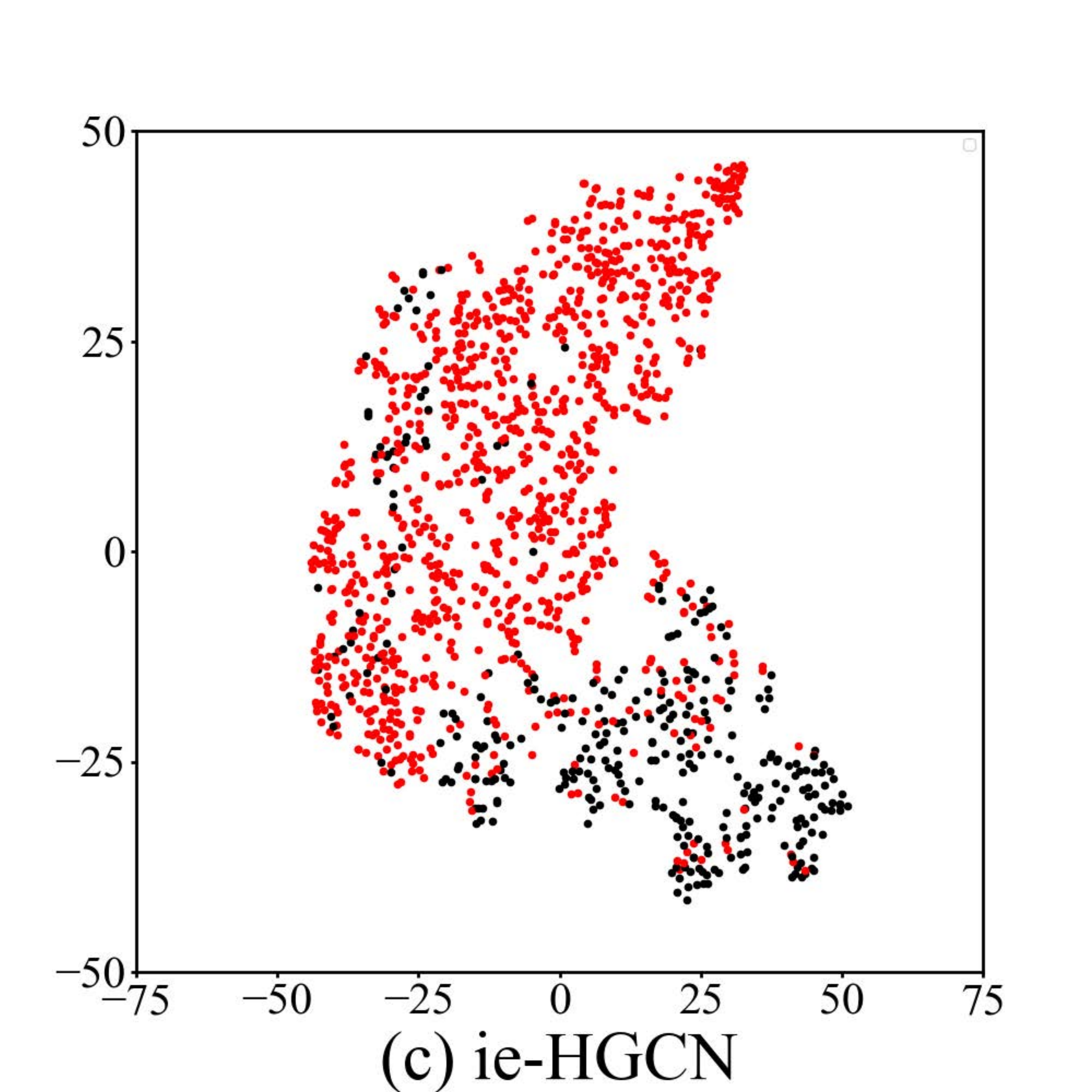}
    \includegraphics[width=0.16\textwidth]{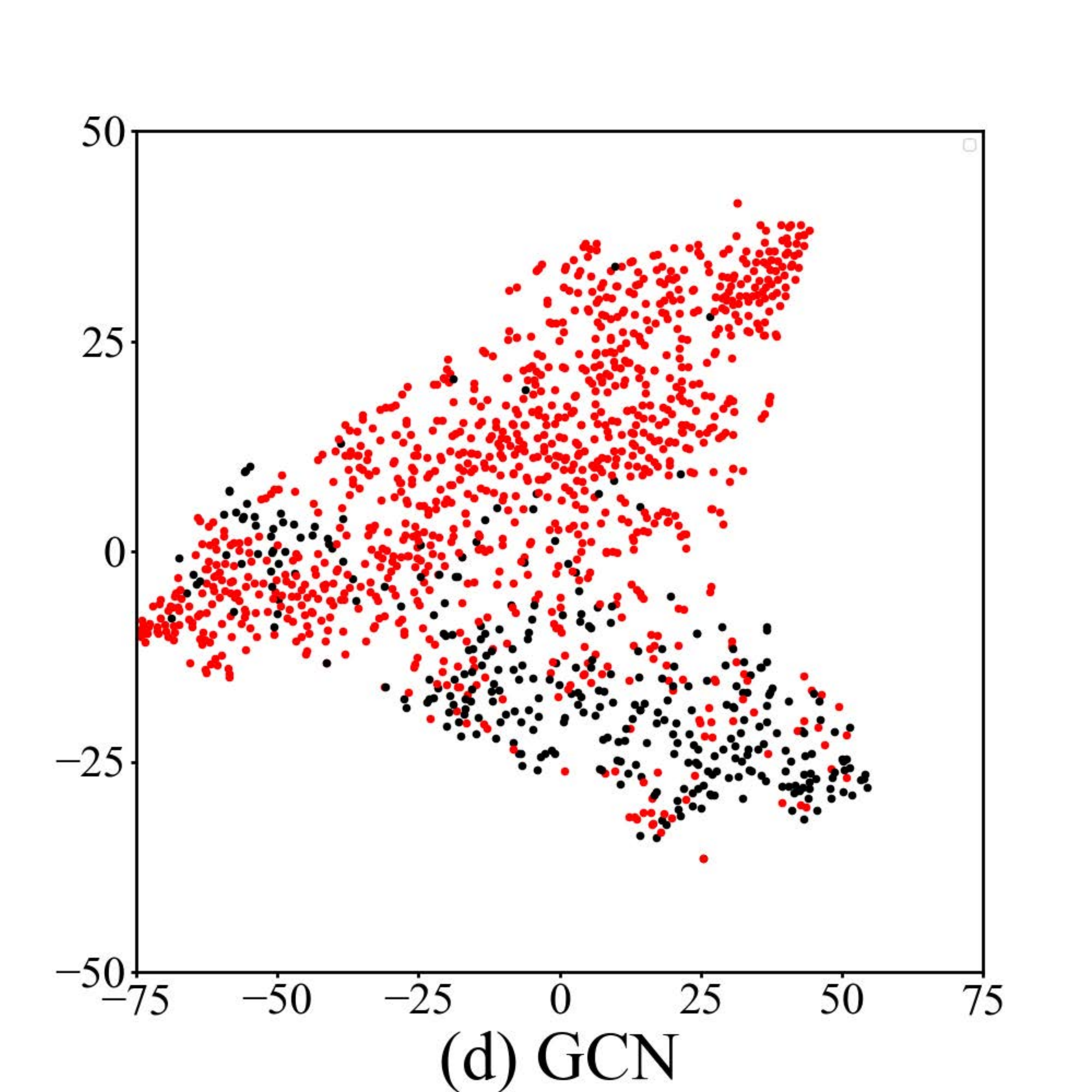}
    \includegraphics[width=0.16\textwidth]{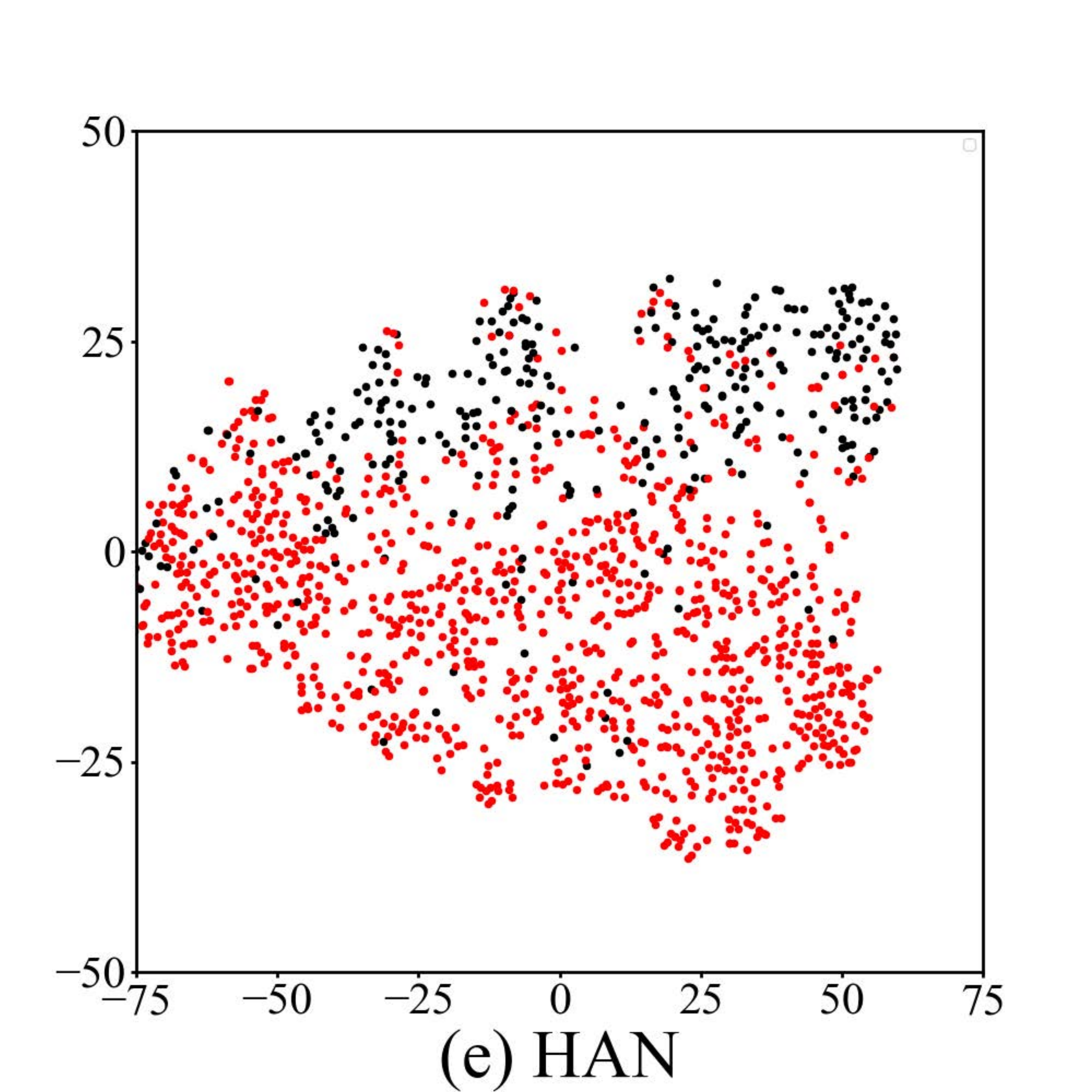}
    \includegraphics[width=0.16\textwidth]{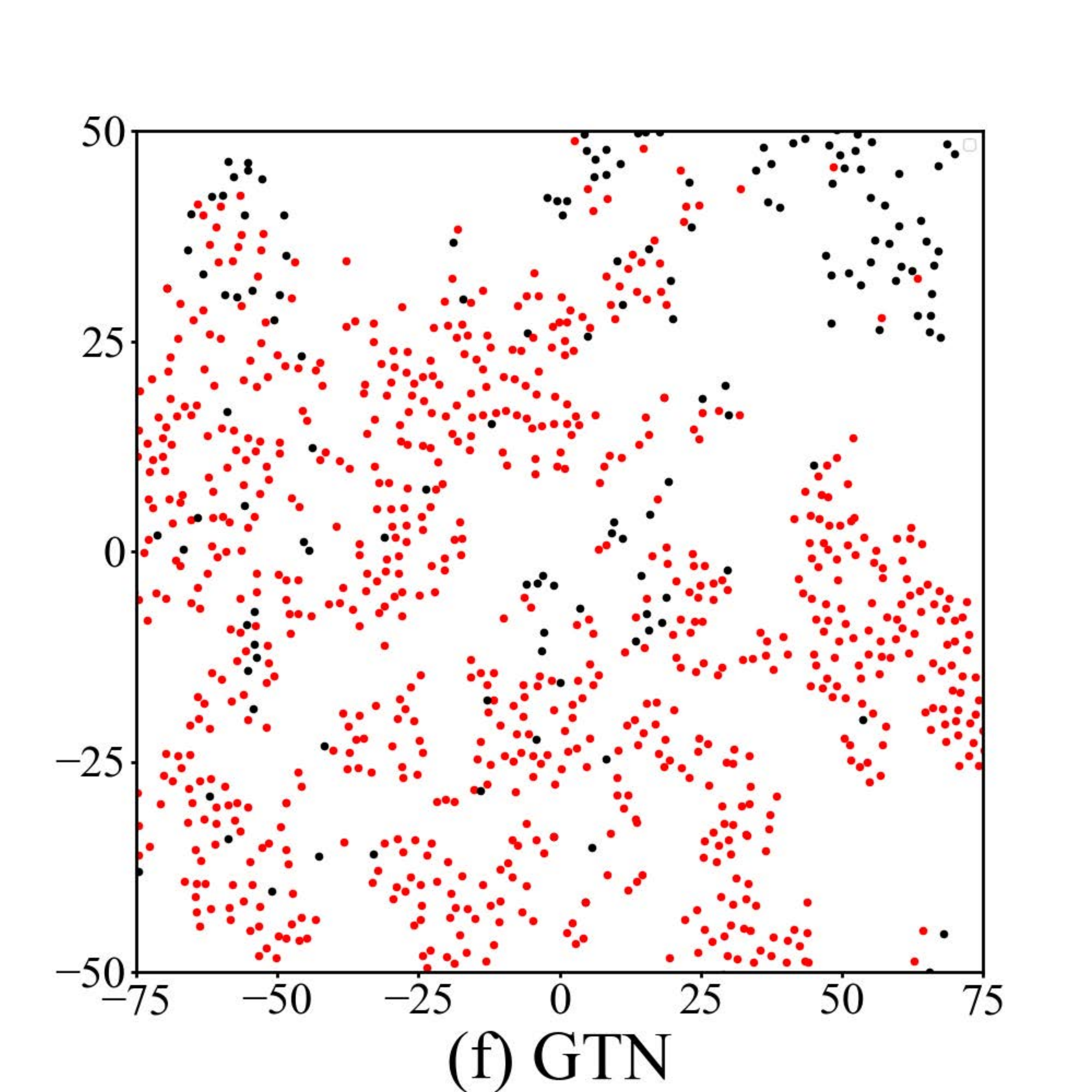}
    
    \caption{Visualization of node representation on OAG1Y. Each point indicates a paper and its color indicates its publication field.}
    \label{fig:Visualizing}
\end{figure*}  

\textbf{Analysis.} (1) Compared with homogeneous GNNs, i.e. GCN and GAT, DHAN achieves significant and consistent performance, which indicates that our proposed model can sufficiently  capture the heterogeneous information from the data. 
(2) Compared with heterogeneous GNNs, i.e., RGCN, HetGNN, GTN and HGT, the proposed model DHAN outperforms all baselines in link prediction tasks on all datasets and indicators. This is mainly because our model is specially designed for bi-typed multi-relational graphs. Hence, it can sufficiently utilize interactions between two types of nodes, which can not be well captured by general heterogeneous graph neural networks. 
Besides, the proposed model also achieves comparable results in classification tasks on most of datasets and indicators. The observation confirms that our model is able to distinguish different relations delicately by utilizing the hierarchical mechanism.
(3) Compared with the conventional hierarchical attention model HGConv and ie-HGCN, our model performs better on all tasks in all datasets. Our model takes advantage of the two typical hierarchical models by fusing relation global information and local information. To be more specific, HAN proposed to aggregate different types of relation information with same global importance, which overlooks heterogeneity of different nodes. HGConv aggregates relation information with heterogeneous weight related to different nodes, which neglects common information that a type of relation has among all nodes. In contrast, our model overcomes their limits by incorporating both the merits of relation global information as well as local information.
(4) In sum, we believe the better performance is due to the better design of our model. First, DHAN can gain improvements via taking both the node intra-class and inter-class attention into consideration. Second, our model also uses an efficient hierarchical attention mechanism to encode the graph.




\subsection{Node Clustering}
We conduct node clustering based on the paper-venue task on three datasets. Here, we first get node representations via feed forward of each GNN. We them apply K-Means to implement node clustering and evaluate the performance using NMI and ARI based on their ground truth and predicted categories. 
Since the results tend to be affected by initial centroids, to make performance more stable, we repeat the former process 10 times and report average results in Table \ref{cluster-table}. 
Experiments results show that out model outperforms all baselines, e.g. on \textit{OAG1Y}, DHAN outperforms the SOTA model ie-HGCN with a margin as large as 0.0277 on ARI.
The results demonstrate the superiority of the learned node representations.


\subsection{Ablation Study.}
To evaluate the contribution of different model components of DHAN, we conduct an ablation study. We generate variants of DHAN by adjusting the use of its model components and comparing their performance on three tasks on \textit{OAG1Y}. The three ablated variants are as follows: 
(1) \textbf{DHAN w/o dual operation}, which does not distinguish the node intra-class and inter-class relation, and only takes one hierarchical attention. 
(2) \textbf{DHAN w/o hierarchical architecture}, which deletes hierarchical architecture in both intra-class and inter-class encoders. 
(3) \textbf{DHAN w/o global attention}, which deletes the relation global attention.

Figure \ref{fig:ablation-study} shows the results of the variants on all three datasets, from which we can observe that removing either dual operation or hierarchical architecture will lead to performance decreasing. 
Specifically, the proposed model DHAN significantly  outperforms \textbf{DHAN w/o dual operation}, which confirms the benefits of the dual mechanism. Thus, we highlight the importance of designing a specific model architecture on the bi-typed graphs rather than a general heterogeneous graph model. 
Compared with \textbf{DHAN w/o global attention} and \textbf{DHAN w/o hierarchical architecture}, we can find that fusing both global information and local information makes a great contribution to the performance of DHAN. Moreover, we could also observe that \textbf{DHAN w/o global attention} always performs better than \textbf{DHAN w/o hierarchical architecture}, which is in line with the fact that \textbf{DHAN w/o hierarchical architecture} is also a simplified version of \textbf{DHAN w/o global attention} removing local attention mechanism. 



\subsection{Visualization.}
To make a more intuitive comparison, we project the representations of paper nodes into two-dimensional space by t-SNE \cite{van2008visualizing}. The node representations are learned on \textit{OAG1Y} based on PF\_L1 tasks.
We randomly choose two fields that no papers belongs to both. 
The color indicates the publishing field of the papers in Figure \ref{fig:Visualizing}. The less mixed areas the better. 
We can observe that our model DHAN performs best in visualization as there are more distinct boundaries and fewer mixed nodes. Besides, we also find that those hierarchical heterogeneous models, i.e., ie-HGCN, HGConv, perform better than general heterogeneous graph models, i.e., HAN, GTN.

\subsection{Variant Analysis}
\begin{figure}[htb]
    \centering
    \includegraphics[width=0.5\textwidth]{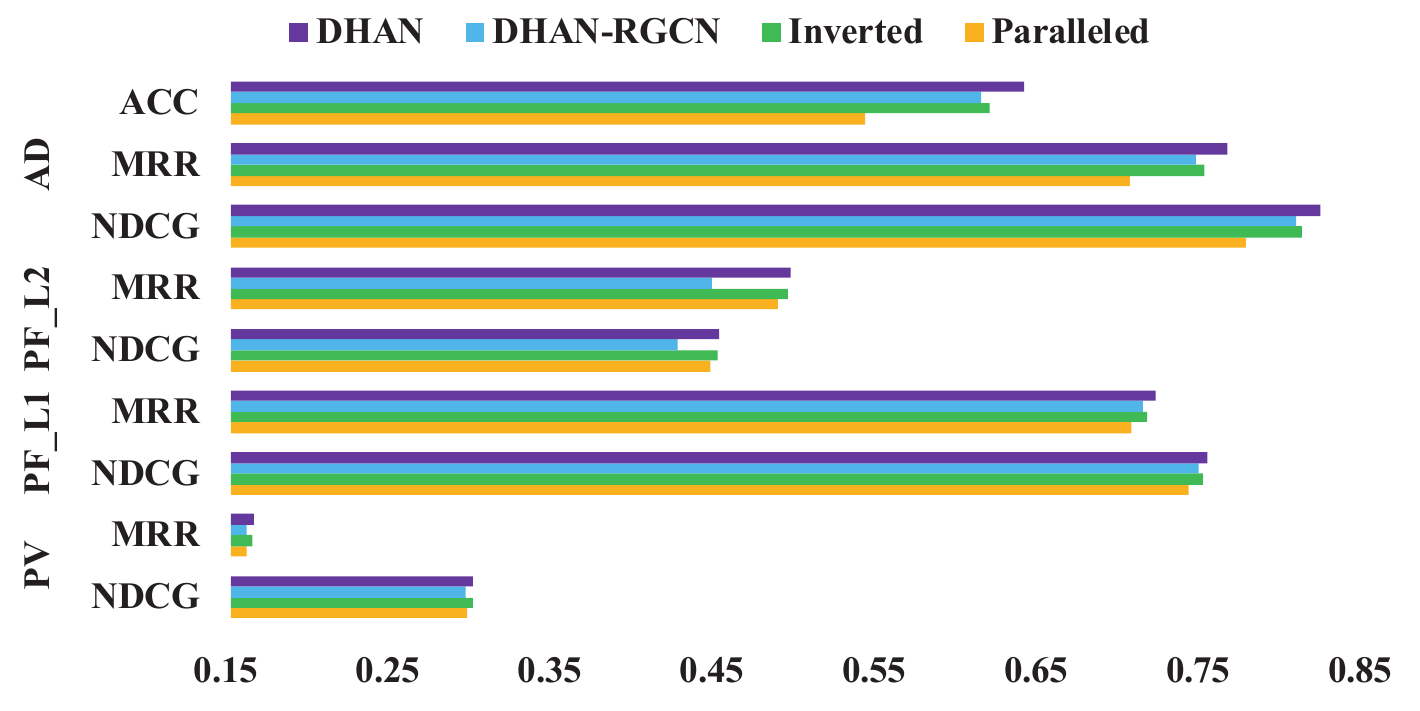}
    \caption{Variant analsis of DHAN on \textit{OAG1Y}.}
    \label{fig:Variants}
\end{figure}

We conduct variant analysis of DHAN on \textit{OAG1Y} with four tasks to show the effectiveness of its architecture. 
(1) \textbf{DAHN-RGCN} substitutes the proposed hierarchical attention mechanism with RGCN and keeps model structure unchanged.
(2) \textbf{Inverted Architecture} firstly implements inter-class hierarchical aggregation and then applies intra-class hierarchical aggregation. 
(3) \textbf{Parallel Architecture} conducts intra-class and inter-class hierarchical aggregation simultaneously and concatenates the updated representation of two types of nodes respectively. 
The results are shown in Figure \ref{fig:Variants}, from which we can observe that all the variants perform worse than DHAN. \textbf{DHAN-RGCN} utilizes RGCN rather than our hierarchical module to aggregate different types of relation information, which thus leads to a performance decrease. The proposed DHAN performs better than both \textbf{Inverted Architecture} and \textbf{Parallel Architecture}, which demonstrates our model structure is a more efficient architecture,i.e., first conducting intra-class relation aggregating then implementing inter-class relation.

\subsection{Interpretability of the Hierarchical Attention}
\begin{figure}[htb]
    \centering
    \includegraphics[width=0.3\textwidth]{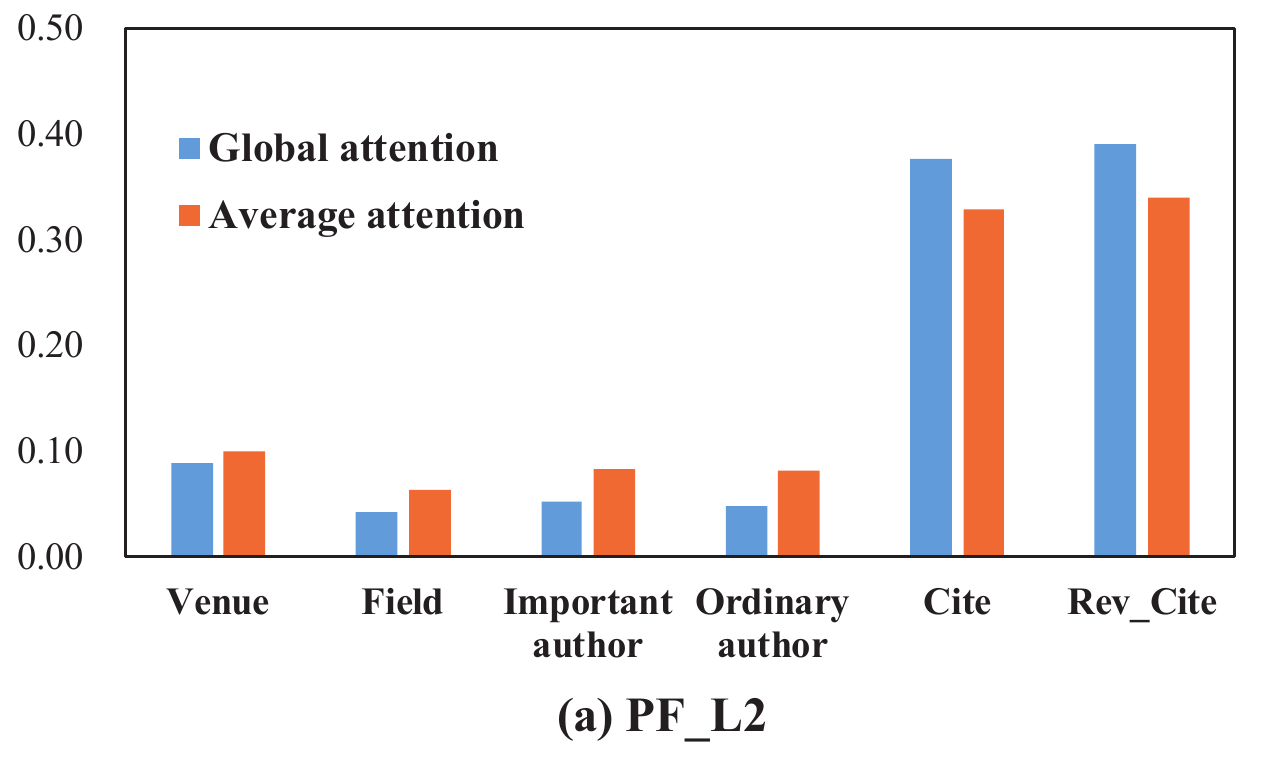}\\
    \includegraphics[width=0.3\textwidth]{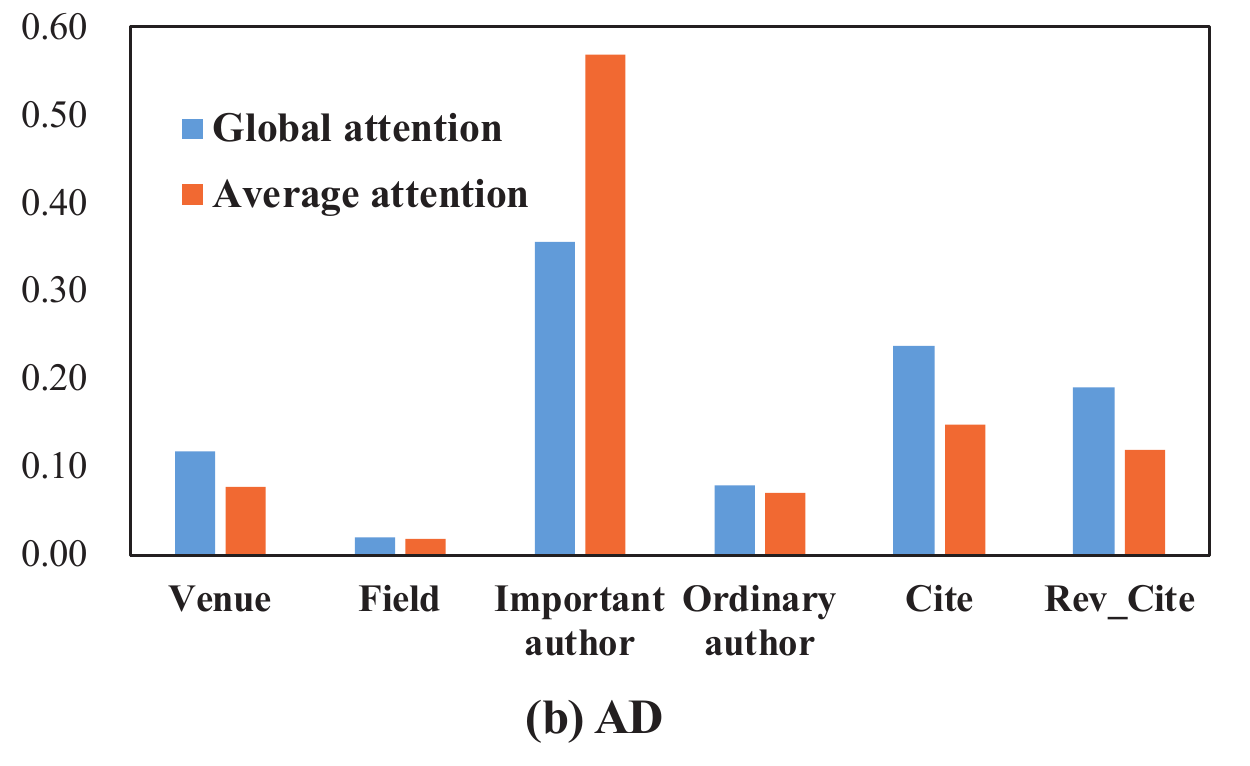}
    \caption{The presentation of the learned attention scores of DHAN. }
    \label{fig:case-study}
\end{figure}

To show the proper interpretability of DHAN, we show the learned attention scores of our model in Figure \ref{fig:case-study}. The global attention is learned parameter for different relations, and the average attention is the finally averaged attention score which is calculated as  the sum of global attention score and heterogeneous attention score of all nodes. Here, we show the results of PF\_L2 task and AD task on \textit{OAG10Y}.

Specifically, we can observe from Figure \ref{fig:case-study} (a) that the learned global attention score of  relation \textit{cite} and  \textit{rev\_cite} gain more weight than other relations in PF\_L2 task. This is in line with the fact that those papers which are either cited by or cite target paper contribute much more than other related papers to the target paper while performing paper field tasks. 
Besides, the ``\textit{is\_important\_author\_of}" and ``\textit{is\_ordinary\_author\_of}" relationships obtain more significant weight than the ``\textit{is\_same\_venue\_of}" and ``\textit{is\_same\_field\_of}" relationships, which is also in line with intuition. 
Moreover, the ``\textit{is\_important\_author\_of}" relationship acquires a bit more considerable weight than ``\textit{is\_ordinary\_author\_of}", which confirms the interpretability of our model again. 
A similar conclusion on AD task is shown in \ref{fig:case-study} (b). 
However, different from Figure \ref{fig:case-study} (a), the global attention weight of ``\textit{is\_important\_author\_of}" is the largest one among all relations, which denotes that papers with same important author have much more influence than other related papers in the author disambiguation task. 
This is mainly because that the author disambiguation task cares more about relations between authors and papers, which is also in line with our intuition. 
Above all, we can find that the average attention score of each relation is significantly different from global attention weight. Actually, in PF\_L2 task, the average attention score of \textit{cite} relation and corresponding standard variance are 0.3293 and 0.0124. In AD task, the average attention score of ``\textit{is\_important\_author\_of}" and corresponding standard variance is 0.5682 and 0.0960. The former two facts demonstrate the necessity of combining both global information and local information for aggregation.

\begin{figure}[htb]
    \centering
    \includegraphics[width=0.5\textwidth]{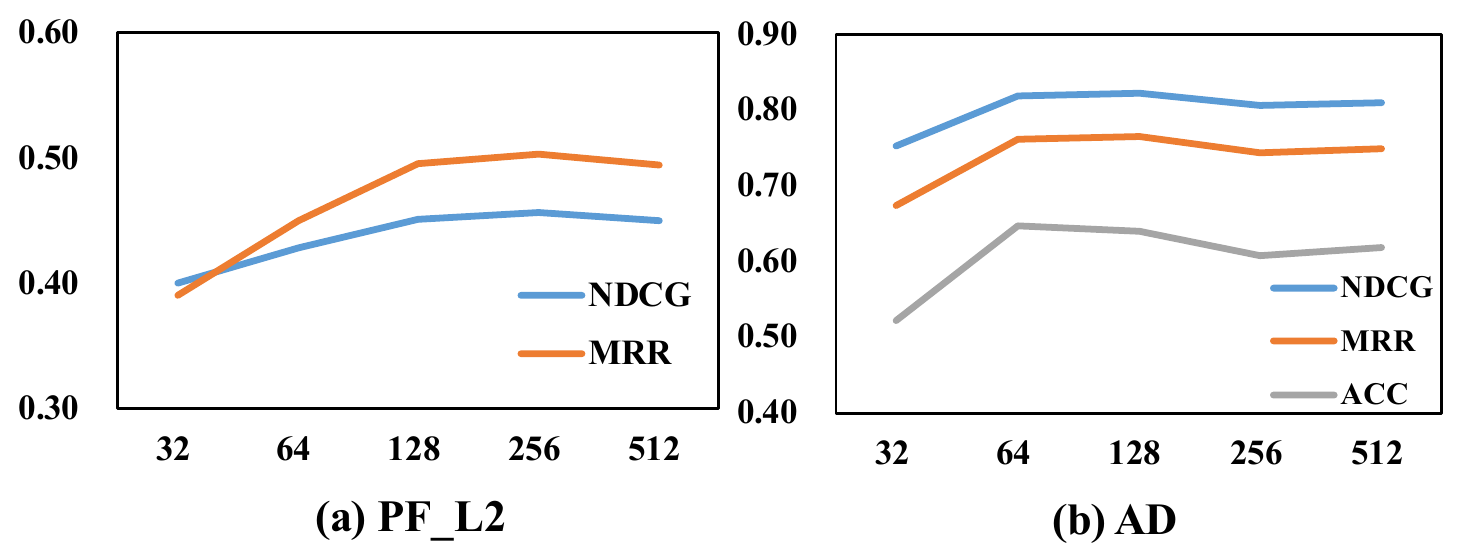} 
    \caption{Parameter sensitivity of DHAN on PF\_L2 and AD task with different dimensions in \textit{OAG1Y}.}
    \label{fig:parameter}
\end{figure}
\subsection{Parameter Analysis}
The hyper-parameter plays an vital role in model performance, and one of the most essential hyper-parameter is the dimension of representations. We conduct parameter analysis in the PF\_L2 and AD task on the \textit{OAG1Y} dataset. The results are shown in Figure \ref{fig:parameter}, from which we can observe that the proposed model reaches its best performance when the dimension of output representation is set as 128. Specifically, the performance first rises with the dimension increasing and then reaches its optimal state since the model needs larger dimension to embody rich information. After that, the performance decreases as a result of overfitting.

\section{Conclusion and Future Work}
\label{section-future_work}
In this paper, we focus on how to learn node efficient representations on bi-typed multi-relational heterogeneous graph. To this end, we propose a novel Dual Hierarchical Attention Networks (DHAN). To the best of our knowledge, we are the first attempt to deal with this task. Specifically, DHAN contains intra-class and inter-class attention-based encoders which enables DHAN to sufficiently leverage not only the node intra-class neighboring information but also the inter-class neighboring information in BMHG. 
Moreover, we adopt a newly proposed hierarchical mechanism to to sufficiently model node multi-relational information in BMHG.
By doing so, the proposed dual hierarchical attention operations enable our model to fully capture the complex structures of the BMHGs. 
We conduct extensive experiments on various tasks against the state-of-the-arts, which sufficiently confirms the capability of DHAN in learning node comprehensive representations in BMHGs. 
Interesting future work directions include generalizing DHAN to other BMHG-based applications.



\ifCLASSOPTIONcompsoc
  \section*{Acknowledgments}
\else
  \section*{Acknowledgment}
\fi

The authors would like to thank all anonymous reviewers in advance.
This research has been partially supported by grants from the National Natural Science Foundation of China under Grant No. 71725001, 71910107002, 61906159, 62176014, U1836206, 71671141, 71873108, 62072379, the State key R \& D Program of China under Grant No. 2020YFC0832702, the major project of the National Social Science Foundation of China under Grant No. 19ZDA092.
and the Financial Intelligence and Financial Engineering Key Laboratory of Sichuan Province.

\ifCLASSOPTIONcaptionsoff
  \newpage
\fi



\bibliographystyle{IEEEtran}
%




\bibliography{sample-base}



\begin{IEEEbiography}[{\includegraphics[width=1in,height=1.25in,clip,keepaspectratio]{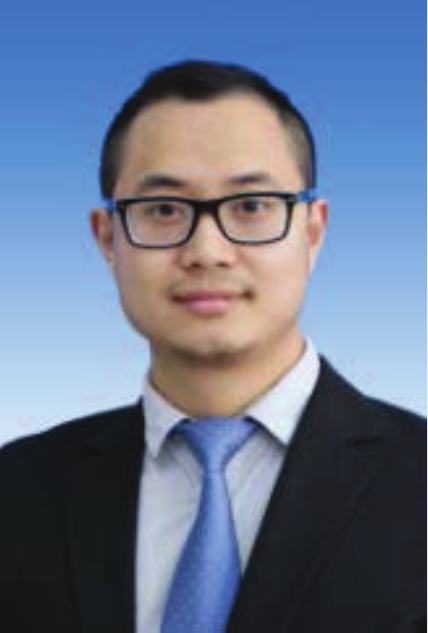}}]{Yu Zhao}
received the B.S. degree from Southwest Jiaotong University in 2006, and the M.S. and Ph.D. degrees from the Beijing University of Posts and Telecommunications in 2011 and 2017, respectively. He is currently an Associate Professor at Southwestern University of Finance and Economics. His current research interests include machine learning, natural language processing, knowledge graph, Fintech. He has authored more than 20 papers in top journals and conferences including IEEE TKDE, IEEE TNNLS, ACL.
\end{IEEEbiography}

\begin{IEEEbiography}[{\includegraphics[width=1in,height=1.25in,clip,keepaspectratio]{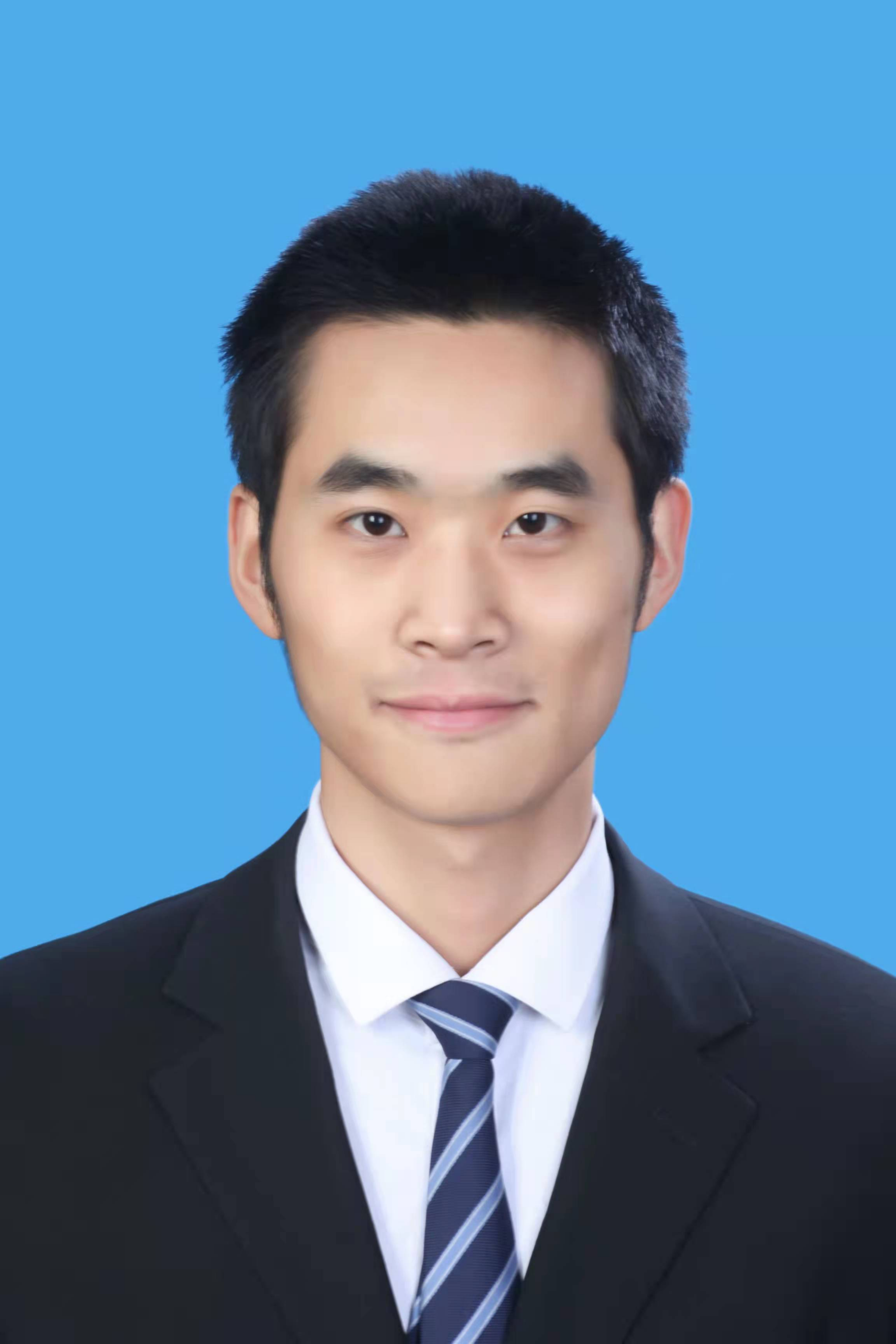}}]{Shaopeng Wei} received the B.S. degree from Huazhong Agricultural University in 2019, and now is a Ph.D student in Southwestern University of Finance and Economics. His research interests include graph learning and relevant applications in recommendation system and Fintech.
\end{IEEEbiography}

\begin{IEEEbiography}[{\includegraphics[width=1in,height=1.25in,clip,keepaspectratio]{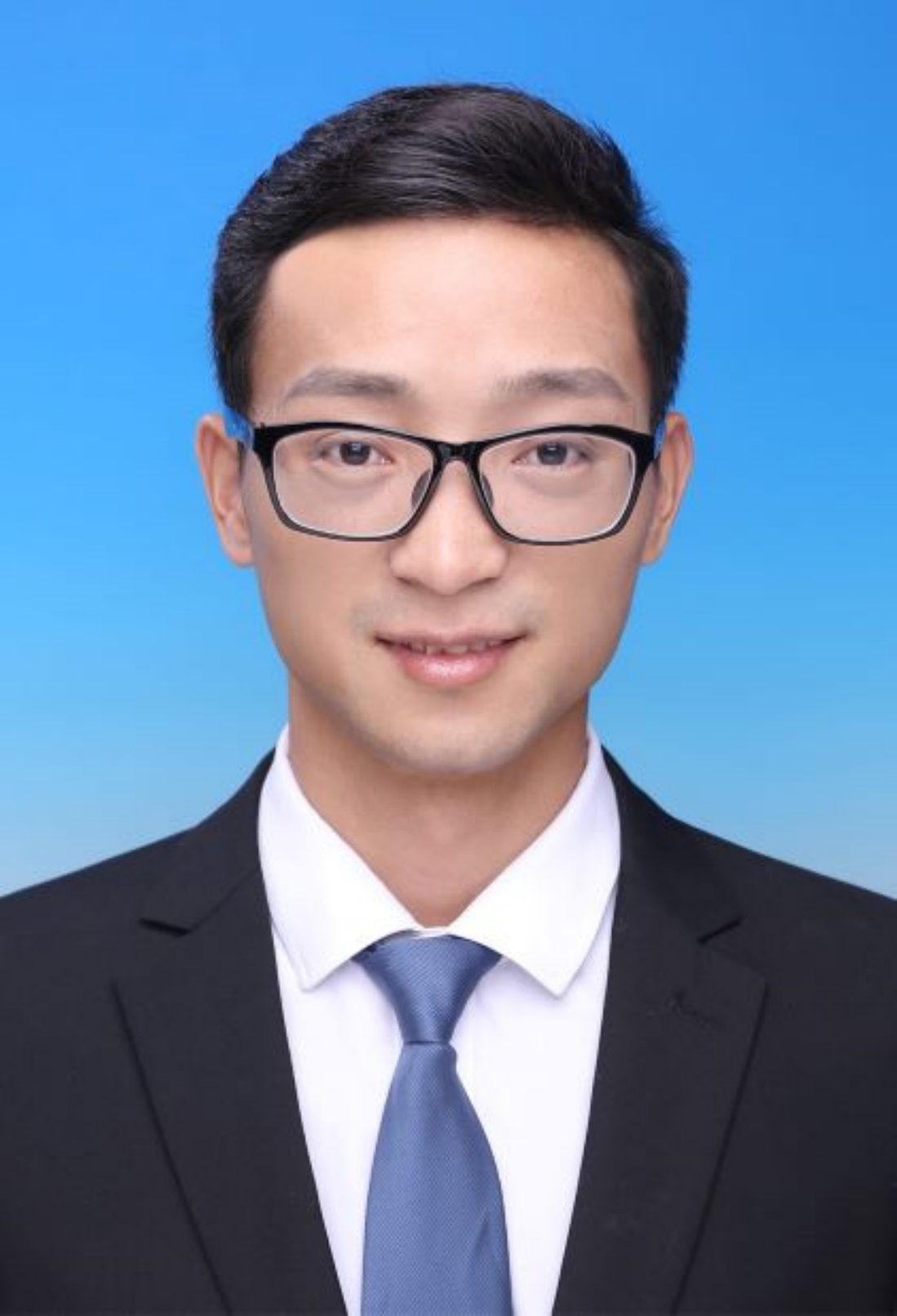}}]{Huaming Du} received his M.S. degree from China University of Petroleum, Beijing, China, in 2018. He is currently pursuing the Ph.D. degree in Southwestern University of Finance and Economics, Chengdu, China. His research interests include Fintech, reinforcement learning, and graph representation learning.
\end{IEEEbiography}

\begin{IEEEbiography}[{\includegraphics[width=1in,height=1.25in,clip,keepaspectratio]{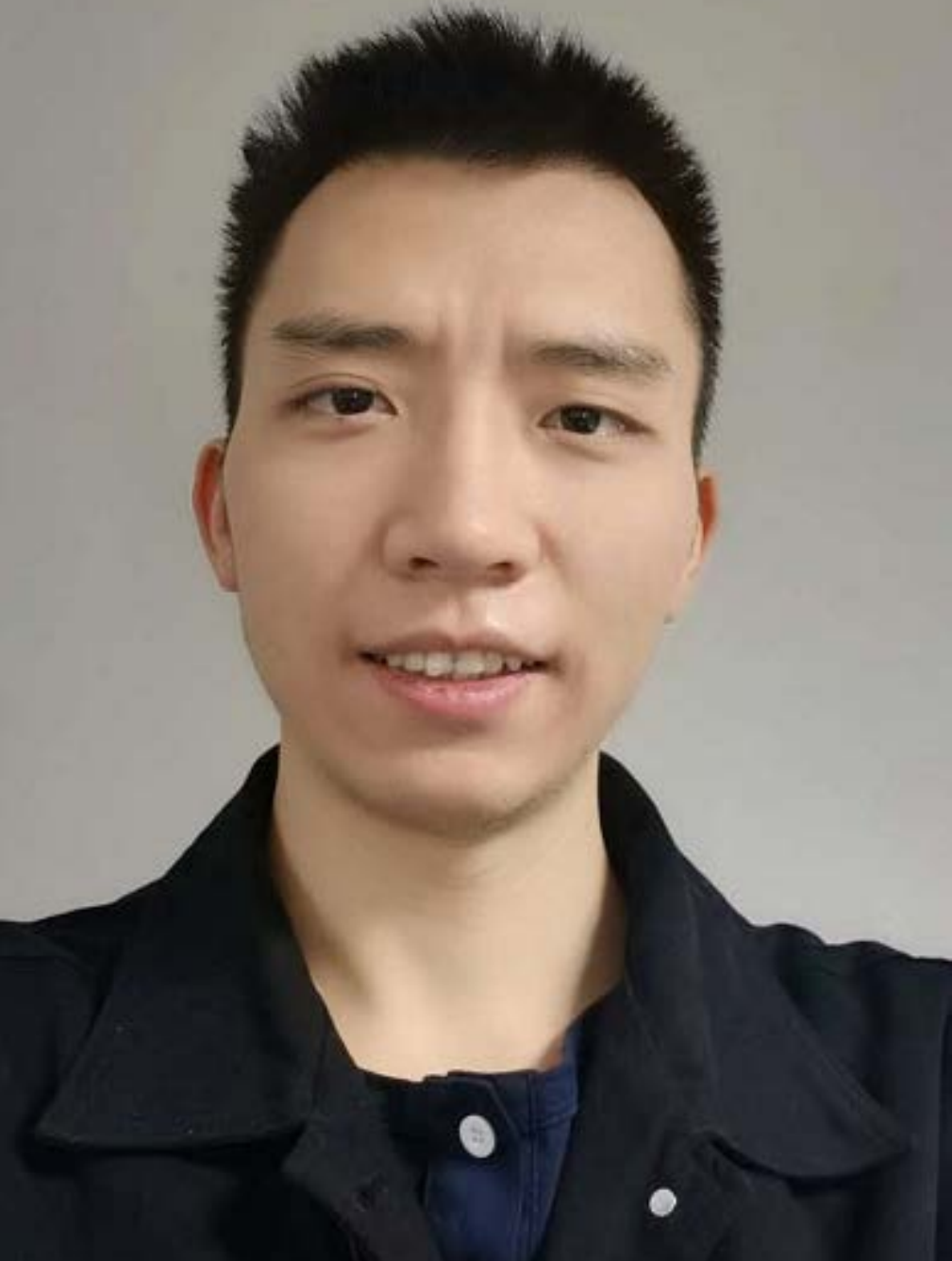}}]{Xingyan Chen}
	received the Ph. D degree in computer technology from Beijing University of Posts and Telecommunications (BUPT), in 2021. 
	He is currently a lecturer with the School of Economic Information Engineering, Southwestern University of Finance and Economics, Chengdu.
	He has published papers in well-archived international journals and proceedings, such as the \textsc{IEEE Transactions on Mobile Computing}, \textsc{IEEE Transactions on Circuits and Systems for Video Technology}, \textsc{IEEE Transactions on Industrial Informatics}, and \textsc{IEEE INFOCOM} etc. 
	His research interests include Multimedia Communications, Multi-agent Reinforcement Learning and Stochastic Optimization.
\end{IEEEbiography}

\begin{IEEEbiography}[{\includegraphics[width=1in,height=1.25in,clip,keepaspectratio]{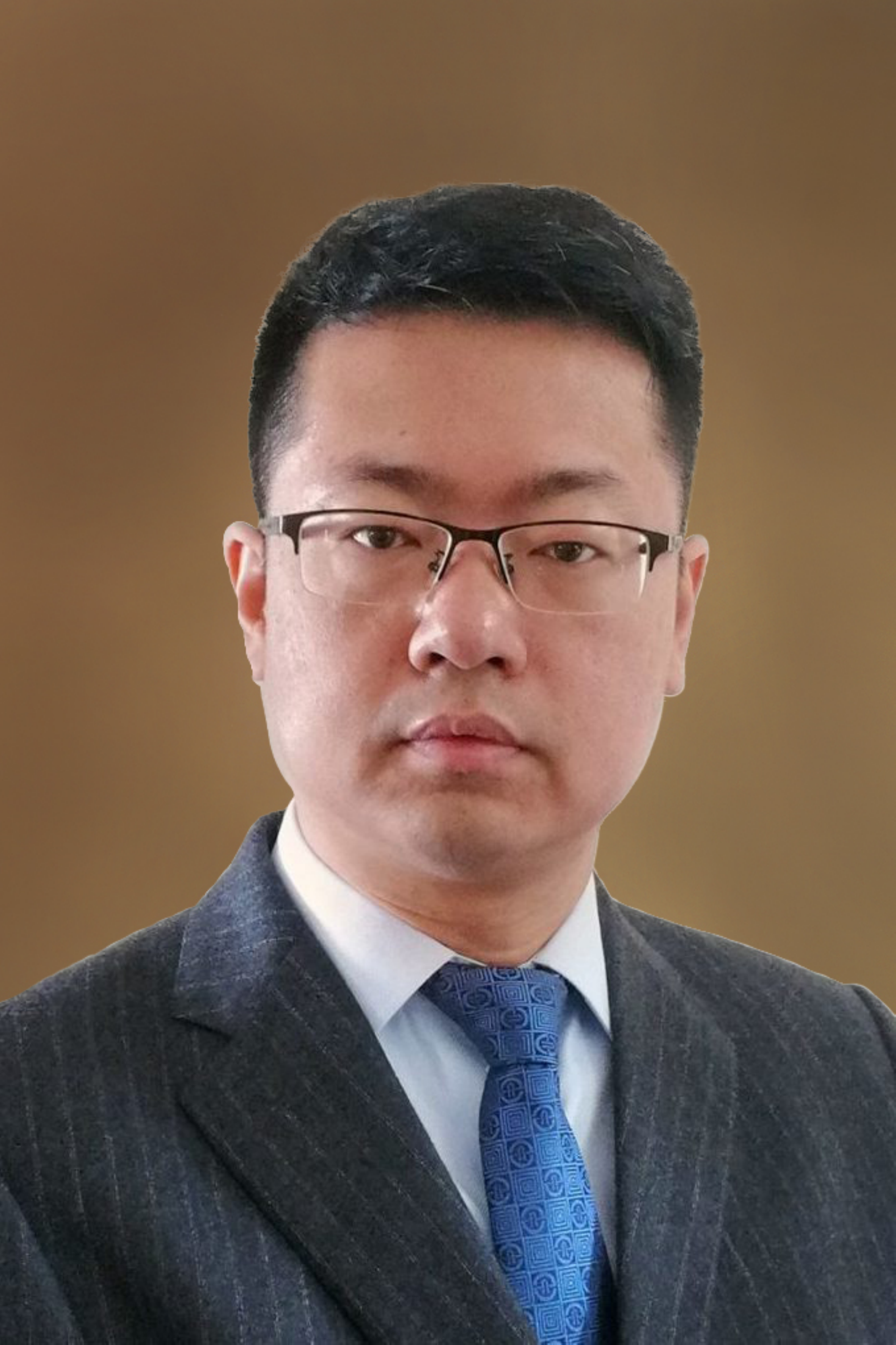}}]{Qing Li} received his PhD degree from Kumoh National Institute of Technology in February of 2005, Korea, and his M.S. and B.S. degrees from Harbin Engineering University, China. He is a postdoctoral researcher at Arizona State University and the Information \& Communications University of Korea. He is a professor at Southwestern University of Finance and Economics, China. His research interests include natural language processing, FinTech. He has published more than 70 papers in the prestigious refereed conferences and journals, such as IEEE TKDE, ACM TOIS, AAAI, SIGIR, ACL, WWW, etc.
\end{IEEEbiography}

\begin{IEEEbiography}[{\includegraphics[width=1in,height=1.25in,clip,keepaspectratio]{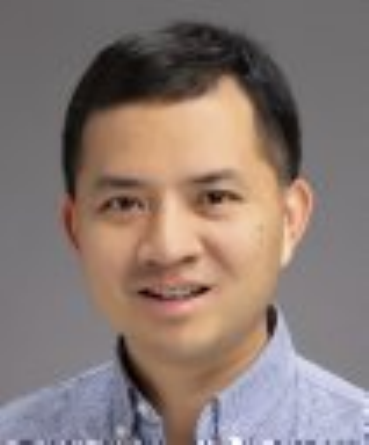}}]{Fuzhen Zhuang} received the PhD degree in computer science from the Institute of Computing Technology, Chinese Academy of Sciences. He is currently a full Professor in Institute of Artificial Intelligence, Beihang University., Beijing 100191, China. His research interests include Machine Learning and Data Mining, including Transfer Learning, Multi-task Learning, Multi-view Learning and Recommendation Systems. He has published more than 100 papers in the prestigious refereed conferences and journals, such as KDD, WWW, SIGIR, ICDE, IJCAI, AAAI, EMNLP, Nature Communications, IEEE TKDE, ACM TKDD, IEEE T-CYB, IEEE TNNLS, ACM TIST, etc.
\end{IEEEbiography}

\begin{IEEEbiography}[{\includegraphics[width=1in,height=1.25in,clip,keepaspectratio]{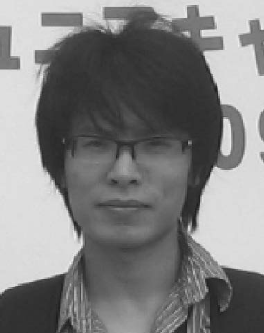}}]{Ji Liu}
received the B.S. degree from the University of Science and Technology of China, Hefei, China, in 2005, the master’s degree from Arizona State University, Tempe, AZ, USA, in 2010, and the Ph.D. degree from the University of Wisconsin–Madison, Madison, WI, USA, in 2014. He is currently an Assistant Professor of computer science, electrical and computer engineering with the Goergen Institute for Data Science, University of Rochester (UR), Rochester, NY, USA, where he created the Machine Learning and Optimization Group. 
He has authored more than 70 papers in top journals and conferences including JMLR, TPAMI, TNNLS, TKDD, NIPS, ICML, SIGKDD, ICCV, and CVPR. Dr. Liu was a recipient of the Award of Best Paper Honorable Mention at SIGKDD 2010, the Award of Best Student Paper Award at UAI 2015, and the IBM Faculty Award. He is named one MIT technology review’s “35 innovators under 35 in China.”
\end{IEEEbiography}


\begin{IEEEbiography}[{\includegraphics[width=1in,height=1.25in,clip,keepaspectratio]{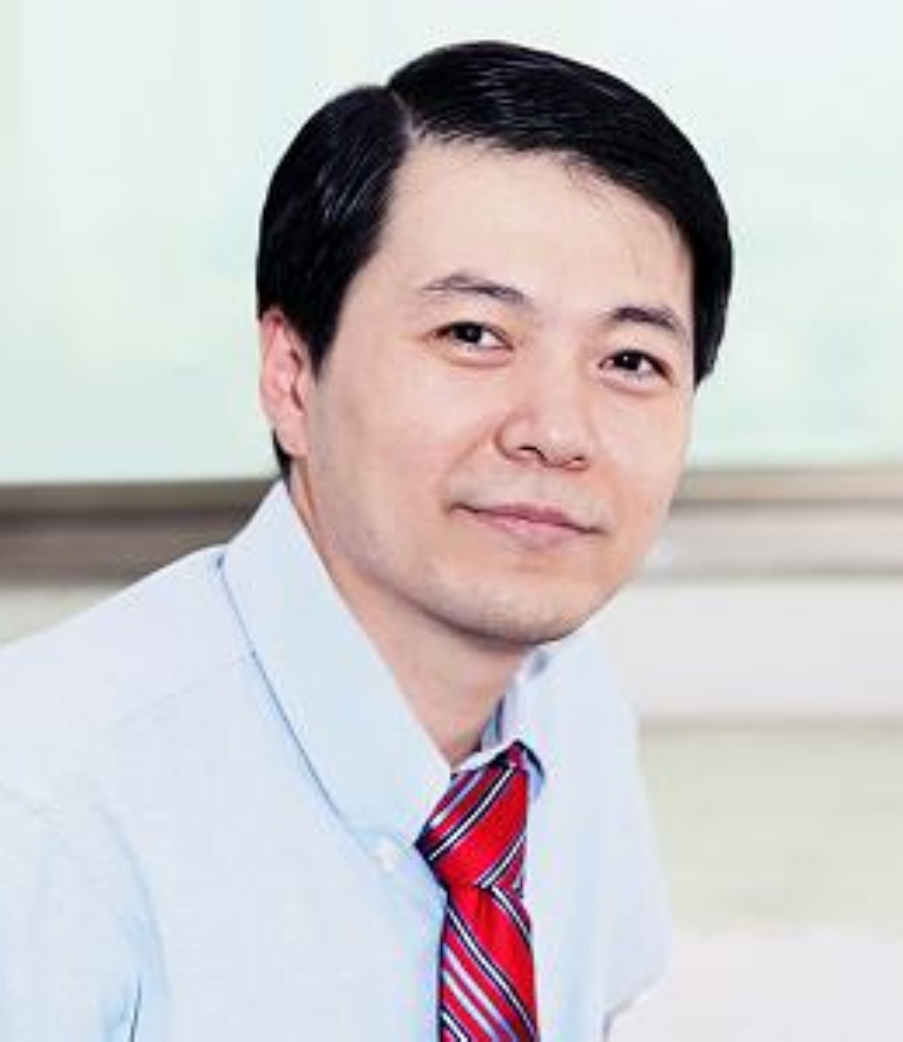}}]{Gang Kou} is a Distinguished Professor of Chang Jiang Scholars Program in Southwestern University of Finance and Economics, managing editor of International Journal of Information Technology \& Decision Making (SCI) and managing editor-in-chief of Financial Innovation (SSCI). He is also editors for other journals, such as: Decision Support Systems, and European Journal of Operational Research. Previously, he was a professor of School of Management and Economics, University of Electronic Science and Technology of China, and a research scientist in Thomson Co., R \& D. He received his Ph.D. in Information Technology from the College of Information Science \& Technology, Univ. of Nebraska at Omaha; Master degree in Dept of Computer Science, Univ. of Nebraska at Omaha; and B.S. degree in Department of Physics, Tsinghua University, China. He has published more than 100 papers in various peer-reviewed journals. Gang Kou’s h-index is 57 and his papers have been cited for more than 10000 times. He is listed as the Highly Cited Researcher by Clarivate Analytics (Web of Science).
\end{IEEEbiography}









\end{document}